\documentclass{article} % For LaTeX2e
\usepackage{iclr2026_conference,times}

\usepackage{amsmath,amsfonts,bm}

\def\eqref#1{equation~\ref{#1}}
\def\1{\bm{1}}

\DeclareMathAlphabet{\mathsfit}{\encodingdefault}{\sfdefault}{m}{sl}
\SetMathAlphabet{\mathsfit}{bold}{\encodingdefault}{\sfdefault}{bx}{n}

\definecolor{softblue}{rgb}{0.21,0.49,0.74}
\usepackage[breaklinks,colorlinks,allcolors=softblue]{hyperref}
\usepackage{url}
\usepackage{amsmath}
\usepackage{amsthm}
\usepackage{booktabs}
\usepackage{multirow}
\usepackage{graphicx}
\usepackage{xcolor}
\usepackage{colortbl}
\usepackage{enumitem}
\usepackage{wrapfig}
\usepackage{subfigure}
\usepackage{amssymb}

\usepackage[breakable]{tcolorbox}
\usepackage{lipsum} % 仅用于生成示例文字
\definecolor{promptbox_bg}{HTML}{F0F8FF}       % 盒子主体的淡蓝色背景
\definecolor{promptbox_frame}{HTML}{5D8DBE}   % 盒子的深蓝色边框
\definecolor{instruction_red}{HTML}{C00000}   % {Here is instruction.} 的红色
\definecolor{down}{HTML}{B1281C}
\definecolor{up}{HTML}{489638}

\newtcolorbox{promptbox}[1]{
    breakable,
    colback = promptbox_bg, % 设置主体背景色
    colframe = promptbox_frame, % 设置边框颜色
    colbacktitle = promptbox_frame, % 设置标题栏背景色
    coltitle = white, % 设置标题文字颜色为白色
    fonttitle = \bfseries, % 设置标题文字为粗体
    title = {#1}, % 盒子的标题，#1 是一个参数，在使用时传入
    boxrule = 1pt, % 边框线的粗细
    arc = 2mm, % 边框的圆角半径
    bottom=1mm
}

\definecolor{blue}{HTML}{3F48CC}
\definecolor{lightblue}{HTML}{F0F8FF}
\newtheorem{theorem}{Theorem}

\title{Harder Is Better: Boosting Mathematical\\Reasoning via Difficulty-Aware GRPO and\\Multi-Aspect Question Reformulation}
\author{
 Yanqi Dai\textsuperscript{1,2}\thanks{Work done during Yanqi Dai’s internship at AMAP, Alibaba Group.},
 Yuxiang Ji\textsuperscript{3},
 Xiao Zhang\textsuperscript{4},
 Yong Wang\textsuperscript{2}\thanks{Project lead: Yong Wang; Corresponding authors: Zhiwu Lu (first) and Yong Wang.},
 Xiangxiang Chu\textsuperscript{2},
 Zhiwu Lu\textsuperscript{1}\footnotemark[2]
\\
 \textsuperscript{1}Gaoling School of Artificial Intelligence, Renmin University of China\\
 \textsuperscript{2}AMAP, Alibaba Group~~
 \textsuperscript{3}Xiamen University~~
 \textsuperscript{4}Dalian University of Technology
\\
}
\iclrfinalcopy % Uncomment for camera-ready version, but NOT for submission.
\begin{document}

\maketitle

\vspace{-0.1in}
\begin{abstract}
\vspace{-0.05in}
Reinforcement Learning with Verifiable Rewards (RLVR) offers a robust mechanism for enhancing mathematical reasoning in large models.
However, we identify a systematic lack of emphasis on more challenging questions in existing methods from both algorithmic and data perspectives, despite their importance for refining underdeveloped capabilities.
Algorithmically, widely used Group Relative Policy Optimization (GRPO) suffers from an implicit imbalance where the magnitude of policy updates is lower for harder questions.
Data-wise, augmentation approaches primarily rephrase questions to enhance diversity without systematically increasing intrinsic difficulty.
To address these issues, we propose a two-dual MathForge framework to improve mathematical reasoning by targeting harder questions from both perspectives, which comprises a Difficulty-Aware Group Policy Optimization (DGPO) algorithm and a Multi-Aspect Question Reformulation (MQR) strategy.
Specifically, DGPO first rectifies the implicit imbalance in GRPO via difficulty-balanced group advantage estimation, and further prioritizes harder questions by difficulty-aware question-level weighting.
Meanwhile, MQR reformulates questions across multiple aspects to increase difficulty while maintaining the original gold answer.
Overall, MathForge forms a synergistic loop: MQR expands the data frontier, and DGPO effectively learns from the augmented data.
Extensive experiments show that MathForge significantly outperforms existing methods on various mathematical reasoning tasks.
The code and augmented data are all available.\footnote{\href{https://github.com/AMAP-ML/MathForge}{https://github.com/AMAP-ML/MathForge}}
\end{abstract}

\vspace{-0.15in}
\section{Introduction}
\vspace{-0.05in}

Recently, large language models (LLMs) have demonstrated remarkable reasoning capabilities, fundamentally altering the landscape of artificial intelligence \citep{jaech2024openai, comanici2025gemini, guo2025deepseek}.
In this context, reinforcement learning with verifiable rewards (RLVR) has been proven as a promising training paradigm \citep{guo2025deepseek, wen2025reinforcement}, especially for enhancing mathematical reasoning.
It adopts rule-based rewards instead of neural reward models, thereby significantly reducing computational overhead and mitigating the risk of reward hacking.

From an algorithmic perspective, the most representative approach to support RLVR is Group Relative Policy Optimization (GRPO) \citep{shao2024deepseekmath}, which estimates relative advantages of a group of responses to the same question.
However, we reveal and mathematically prove a critical limitation in GRPO and its variants: their advantage estimation function introduces an implicit imbalance where the update magnitudes are suppressed for both easier and harder questions and peak for those of moderate difficulty.
The neglect of more challenging yet solvable questions is detrimental to RL training.
Such questions are ideal training material, as they expose the model's incomplete mastery while also offering at least one correct response for targeted improvement.
Therefore, harder questions should be emphasized to focus the model on overcoming its solvable weaknesses, while easier ones necessitate only minimal yet sufficient weighting to prevent forgetting.
\citet{zhang2025grpo} also recognize the importance of question difficulty in GRPO, but their method proposes a complex difficulty-aware advantage reweighting without rectifying the underlying imbalance.

Meanwhile, from a data perspective, traditional augmentation methods for reasoning often generate entirely new question-answer pairs \citep{luo2023wizardmath, li2023mugglemath, li2024common}, but the quality of the answers is difficult to guarantee, especially for competition-level problems.
As for those tailored for RLVR, only \citet{liang2025beyond} explore rephrasing questions while sustaining the original answer to enhance data diversity.
However, the question difficulty dimension still lacks attention.
Recognizing that solving mathematical reasoning problems requires varying skills, we contend that systematically increasing question difficulty by reformulating them to target and challenge these skills is a crucial approach for pushing the model's performance boundaries.

To address these issues, we introduce a two-dual framework termed MathForge to enhance mathematical reasoning by targeting more challenging questions from both algorithmic and data perspectives.
Specifically, MathForge comprises two key components: a Difficulty-Aware Group Policy Optimization (DGPO) algorithm and a Multi-Aspect Question Reformulation (MQR) strategy.
Algorithmically, DGPO first rectifies the implicit imbalance of the update magnitudes in GRPO via difficulty-balanced group advantage estimation, which normalizes group relative advantages by the mean absolute deviation of rewards rather than the standard deviation employed in GRPO.
Furthermore, DGPO prioritizes harder questions using difficulty-aware question-level weighting, where the question difficulty is quantified as the negative mean accuracy calculated across all its corresponding responses.
Data-wise, MQR reformulates the original questions across multiple aspects to increase their difficulty and diversity, including adding story background, introducing abstract terminology, and nesting sub-problems.
A critical constraint is that all reformulations must preserve the original gold answer, so that MQR can maintain the essential mathematical logic of the question and obviate the need for solution regeneration.
Overall, our MathForge creates a powerful synergistic loop, where MQR expands the data frontier and DGPO efficiently learns from these augmented data.

The main contributions of this paper can be summarized as follows:
\begin{enumerate}[leftmargin=12pt, topsep=-4pt, itemsep=0pt, partopsep=0pt]
\item We introduce a Difficulty-Aware Group Policy Optimization (DGPO) algorithm, which rectifies the implicit imbalance of GRPO and further upweights more challenging questions.
\item We propose Multi-Aspect Question Reformulation (MQR), a data augmentation strategy tailored for RLVR, which reformulates questions across multiple aspects to increase their difficulty while preserving the original gold answer.
\item Experiments show that our MathForge markedly outperforms existing methods on various models and mathematical reasoning benchmarks, validating its effectiveness and generalizability.
\end{enumerate}

\vspace{-0.05in}
\section{Preliminaries}
\vspace{-0.07in}

\textbf{Notation.}~~In this paper, an autoregressive language model, parameterized by $\theta$, is treated as a policy model, where $\pi_\theta$ and $\pi_{\theta_\text{old}}$ represent the current and old policies, respectively. 
For a given query $q$ sampled from a question dataset $\mathcal{D}$, multiple responses $\left\{o_i\right\}$ are generated using the old policy $\pi_{\theta_\text{old}}$.
A scalar reward $r_i$ for each query-response pair $(q,o_i)$ is then assigned by a rule-based verifier.
By default, we only use the accuracy reward, $1$ if the response is correct and $0$ otherwise.
In the context of batch processing, $\left\{q_s\right\}$ signifies a batch of queries sampled from the question dataset $\mathcal{D}$, and the corresponding responses and rewards are denoted by $\{o_{si}\}$ and $\{r_{si}\}$, respectively.

\textbf{Group Relative Policy Optimization (GRPO).} GRPO \citep{shao2024deepseekmath} is a variant of Proximal Policy Optimization (PPO) \citep{schulman2017proximal}, which eliminates the critic model, and estimates relative advantages of responses within a group of responses to the same query.
Moreover, \citet{chu2025gpg} and \citet{yu2025dapo} remove the KL divergence and employ a token-level policy gradient loss to enhance the performance of GRPO. These modifications have been experimentally validated and are more commonly used in practice, becoming the default settings in TRL \citep{vonwerra2022trl}.
Specifically, GRPO optimizes the policy model $\pi_\theta$ by maximizing the following objective:
\begin{multline}
\mathcal{J}_\text{GRPO}(\theta)=\mathbb{E}\left[q\sim\mathcal{D},\left\{o_i\right\}^G_{i=1}\sim\pi_{\theta_\text{old}}(\cdot\mid q)\right]\\
\frac{1}{\sum^G_{i=1}\left|o_i\right|}\sum^G_{i=1}\sum^{\left|o_i\right|}_{t=1}\left\{\min\left[I_{it}(\theta)\hat{A}_{\text{GR},i}, \operatorname{clip}\left(I_{it}(\theta),1-\varepsilon,1+\varepsilon\right)\hat{A}_{\text{GR},i}\right]\right\},
\end{multline}
\begin{equation}\label{eq:grae}
\text{where}~I_{it}(\theta)=\frac{\pi_\theta\left(o_{i,t}\mid q,o_{i,<t}\right)}{\pi_{\theta_\text{old}}\left(o_{i,t}\mid q,o_{i,<t}\right)},~\hat{A}_{\text{GR},i}=\frac{r_i-\operatorname{mean}\left(\left\{r_i\right\}^G_{i=1}\right)}{\operatorname{std}\left(\left\{r_i\right\}^G_{i=1}\right)}.
\end{equation}
Here, $I_{it}(\theta)$ denotes the importance sampling ratio of the token $o_{i,t}$, and $\hat{A}_{\text{GR},i}$ signifies the advantage of the response $o_i$ obtained by group relative advantage estimation (GRAE). $G$ is the number of generated responses to each query $q$ (i.e., the group size), and $\varepsilon$ is the clipping range of $I_{it}(\theta)$.

% 原始 GRPO 公式
% \begin{multline}
% \mathcal{J}_\text{GRPO}(\theta)=\mathbb{E}\left[q\sim\mathcal{D},\left\{o_i\right\}^G_{i=1}\sim\pi_{\theta_\text{old}}(\cdot\mid q)\right]\\
% \frac{1}{G}\sum^G_{i=1}\frac{1}{\left|o_i\right|}\sum^{\left|o_i\right|}_{t=1}\left\{\min\left[s_{i,t}(\theta)\hat{A}_{i,t}, \operatorname{clip}\left(s_{i,t}(\theta),1-\varepsilon,1+\varepsilon\right)\hat{A}_{i,t}\right]-\beta\mathbb{D}_\text{KL}\left[\pi_\theta\| \pi_\text{ref}\right]\right\},
% \end{multline}
% \begin{equation}
% \mathbb{D}_\text{KL}\left[\pi_\theta\|\pi_\text{ref}\right]=\frac{\pi_\text{ref}\left(o_{i,t}\mid q,o_{i,<t}\right)}{\pi_\theta\left(o_{i,t}\mid q,o_{i,<t}\right)}-\log\frac{\pi_\text{ref}\left(o_{i,t}\mid q,o_{i,<t}\right)}{\pi_\theta\left(o_{i,t}\mid q,o_{i,<t}\right)}-1,
% \end{equation}

\vspace{-0.05in}
\section{Methodology}
\vspace{-0.05in}

In this section, we introduce the MathForge framework to enhance mathematical reasoning by concentrating on more challenging questions from both algorithmic and data perspectives.
Specifically, it consists of two core components: the Difficulty-Aware Group Policy Optimization (DGPO) algorithm and the Multi-Aspect Question Reformulation (MQR) strategy.
% the Difficulty-Aware Group Policy Optimization (DGPO) algorithm, which builds upon GRPO to implement difficulty-aware balancing and re-weighting.
% the Multi-Aspect Question Reformulation (MQR) approach, which increases question difficulty by reformulating it from multiple aspects while preserving the ground-truth answer.

\vspace{-0.05in}
\subsection{Difficulty-Aware Group Policy Optimization}
\vspace{-0.05in}
% DGPO

Although GRPO achieves strong reasoning performance, we mathematically prove that its optimization objective is unbalanced with respect to the update magnitudes for questions with varying difficulties, which primarily stems from its group relative advantage estimation (i.e., $\hat{A}_{\text{GR},i}$ in Equation~\ref{eq:grae}).
This imbalance potentially reduces the extent to which the policy updates for more challenging yet solvable questions.
However, such questions are ideal training material that expose the model's incomplete mastery while also offering at least one correct response for targeted improvement.
Moreover, harder questions may be more complex compositions or reformulations of easier ones, thus mastering harder ones can potentially enhance the model's performance on easier ones.

To resolve this issue, our Difficulty-Aware Group Policy Optimization (DGPO) algorithm first proposes difficulty-balanced group advantage estimation (DGAE) to normalize the update magnitudes across questions.
Secondly, it employs difficulty-aware question-level weighting (DQW) to prioritize more challenging questions further.

Specifically, the optimization objective of DGPO is defined as follows:
\begin{multline}
    \mathcal{J}_\text{DGPO}(\theta)=\mathbb{E}\left[\left\{q_s\right\}^B_{s=1}\sim\mathcal{D},\left\{o_{si}\right\}^G_{i=1}\sim\pi_{\theta_\text{old}}(\cdot\mid q_s)\right]\\
    \frac{1}{\sum^{\textcolor{blue}{B_\text{v}}}_{s=1}\sum^G_{i=1}\left|o_{si}\right|}\sum^{\textcolor{blue}{B_\text{v}}}_{s=1}\textcolor{blue}{\lambda_s}\sum^G_{i=1}\sum^{\left|o_{si}\right|}_{t=1}\left\{\min\left[I_{sit}(\theta)\textcolor{blue}{\hat{A}_{\text{DG},si}},\operatorname{clip}\left(I_{sit}(\theta),1-\varepsilon,1+\varepsilon\right)\textcolor{blue}{\hat{A}_{\text{DG},si}}\right]\right\},
\end{multline}
where $I_{sit}(\theta)$ is the importance sampling ratio of the token $o_{si,t}$, and $\hat{A}_{\text{DG},si}$ is the advantage of the response $o_i$ obtained by DGAE, respectively given by:
\begin{equation}
    I_{sit}(\theta)=\frac{\pi_\theta\left(o_{si,t}\mid q_s,o_{si,<t}\right)}{\pi_{\theta_\text{old}}\left(o_{si,t}\mid q_s,o_{si,<t}\right)},~~\hat{A}_{\text{DG},si}=\frac{r_{si}-\operatorname{mean}\left(\left\{r_{si}\right\}^G_{i=1}\right)}{\operatorname{MAD}\left(\left\{r_{si}\right\}^G_{i=1}\right)},
\end{equation}
\begin{equation}
    \text{where}~\operatorname{MAD}\left(\left\{r_{si}\right\}^G_{i=1}\right) = \frac{1}{G}\sum^G_{i=1}\left|r_{si}-\operatorname{mean}\left(\left\{r_{si}\right\}^G_{i=1}\right)\right|.
\end{equation}
Here, $\operatorname{MAD}(\cdot)$ denotes the mean absolute deviation function.
Furthermore, $\lambda_s$ is the difficulty-aware weight for the query $q_s$ computed by DQW as follows:
\begin{equation}
    \lambda_s = B_\text{v}\cdot\frac{\exp\left(D_s/T\right)}{\sum^{B_\text{v}}_{s=1}\exp\left(D_s/T\right)},~~\text{where}~D_s=-\operatorname{mean}\left(\left\{r_{si}\right\}^G_{i=1}\right).
\end{equation}
Here, $B$ represents the global batch size, and $B_\text{v}$ signifies the number of valid queries in the batch.
A query is considered valid if its $G$ corresponding responses are not uniformly correct or incorrect.
Without loss of generality, we assume that the first $B_\text{v}$ queries in the batch are valid.
The token-level average loss is calculated exclusively on valid queries, a procedure we refer to as valid token-level loss averaging.
This design is inspired by GPG \citep{chu2025gpg} and DAPO \citep{yu2025dapo} and is not a key contribution of DGPO. It aims to prevent sharp gradient fluctuations caused by inconsistent valid token ratios across batches, thereby ensuring training stability, and also serves as the basis for valid query reweighting in the following DQW.

In the following subsections, we will describe the two key techniques of DGPO: difficulty-balanced group advantage estimation and difficulty-aware question-level weighting.

\vspace{-0.05in}
\subsubsection{Difficulty-Balanced Group Advantage Estimation}
\vspace{-0.05in}
% DGAE

Consider a single question $q$ and its corresponding responses $\left\{o_i\right\}^G_{i=1}$, the unclipped policy gradient calculated in GRPO is as follows:
\begin{align}\label{eq:gradient}
    g_\text{GRPO} &= \frac{1}{\sum^G_{i=1}\left|o_i\right|}\sum^G_{i=1}\sum^{\left|o_i\right|}_{t=1}\hat{A}_{\text{GR},i}\nabla_\theta I_{it}(\theta) \nonumber\\
    &= \frac{1}{\sum^G_{i=1}\left|o_i\right|}\sum^G_{i=1}\sum^{\left|o_i\right|}_{t=1}\operatorname{sgn}\left(\hat{A}_{\text{GR},i}\right)\left|\hat{A}_{\text{GR},i}\right|\operatorname{detach}\left(I_{it}(\theta)\right)\nabla_\theta\log\left(\pi_\theta\left(o_{i,t}\mid q,o_{i,<t}\right)\right),
\end{align}
where $\operatorname{sgn}(\cdot)$ is the sign function and $\operatorname{detach}(\cdot)$ is the stop-gradient operator.
The full derivation is provided in Appendix~\ref{sec:appendix_gradient}.
In this equation, $\operatorname{detach}\left(I_{it}(\theta)\right)$ and $\nabla_\theta\log\left(\pi_\theta\left(o_{i,t}\mid q,o_{i,<t}\right)\right)$ respectively represent the importance sampling ratio and likelihood gradient for each token $o_{i,t}$. 
Crucially, $\operatorname{sgn}(\hat{A}_{\text{GR},i})$ indicates whether the policy $\pi_\theta$ should be updated to increase or decrease the probability of generating the response $o_i$, while $|\hat{A}_{\text{GR},i}|$ determines the corresponding update magnitude.
Therefore, the total update magnitude for a single question $q$ can be upper-bounded and well-approximated by the sum of these individual magnitudes across all $G$ responses, i.e., $\sum^G_{i=1}|\hat{A}_{\text{GR},i}|$. 
The complete derivation is provided in Appendix~\ref{sec:appendix_magnitude}.
This magnitude has a closed-form expression, as formalized in the following theorem.
The complete proof is provided in Appendix~\ref{sec:appendix_grae}.

\begin{theorem}[Update Magnitude for a Single Question using GRAE]\label{th:grae}
Given a single question $q$ and its corresponding responses $\left\{o_i\right\}^G_{i=1}$, each query-response pair receives a binary accuracy reward $r_i \in \{0,1\}$, and $p$ represents the accuracy rate, i.e., the proportion for a reward of $1$.
Then, the total update magnitude without clipping for the single question $q$ when using GRAE satisfies:
\begin{equation}
    \sum^G_{i=1}\left|\hat{A}_{\textnormal{GR},i}\right| = \sum^G_{i=1}\left|\frac{r_i-\operatorname{mean}\left(\left\{r_i\right\}^G_{i=1}\right)}{\operatorname{std}\left(\left\{r_i\right\}^G_{i=1}\right)}\right| = 2G\sqrt{p(1-p)},~\textnormal{where}~p=\frac{1}{G}\sum^G_{i=1}r_i.
\end{equation}
This total update magnitude varies with respect to the accuracy rate $p$, reaching its maximum when $p=0.5$ and gradually decreasing as $p$ approaches either $0$ or $1$.
\end{theorem}

Theorem~\ref{th:grae} implies that within a training batch, questions with moderate accuracy rates have a greater influence on the policy update, while easier or harder questions have a smaller impact.
However, we argue that challenging questions, yet have non-zero accuracy rates, warrant greater attention.
These questions are ideal for training because they identify areas of the policy model's incomplete mastery while providing at least one correct response for targeted learning.
Consequently, to counteract the inherent imbalance of GRAE, we develop a novel difficulty-balanced group advantage estimation (DGAE) strategy.
Specifically, the advantage function of DGAE is defined as follows:
\begin{equation}
    \hat{A}_{\text{DG},i} = \frac{r_{i}-\operatorname{mean}\left(\left\{r_{i}\right\}^G_{i=1}\right)}{\operatorname{MAD}\left(\left\{r_i\right\}^G_{i=1}\right)},~~\text{where}~\operatorname{MAD}\left(\left\{r_{i}\right\}^G_{i=1}\right) = \frac{1}{G}\sum^G_{i=1}\left|r_{i}-\operatorname{mean}\left(\left\{r_{i}\right\}^G_{i=1}\right)\right|.
\end{equation}
Here, the denominator $\operatorname{MAD}(\cdot)$ is the mean absolute deviation of rewards across all $G$ responses.
This normalization ensures that the total update magnitude for a single question is a constant value, as formalized in the following theorem.
The complete proof is provided in Appendix~\ref{sec:appendix_dgae}.

\begin{theorem}[Update Magnitude for a Single Question using DGAE]\label{th:dgae}
Given a single question $q$ and its corresponding responses $\left\{o_i\right\}^G_{i=1}$, each query-response pair receives a reward $r_i$.
Then, the total update magnitude without clipping for the single question $q$ when using DGAE satisfies:
\begin{equation}
    \sum^G_{i=1}\left|\hat{A}_{\textnormal{DG},i}\right| = \sum^G_{i=1}\left|\frac{r_{i}-\operatorname{mean}\left(\left\{r_{i}\right\}^G_{i=1}\right)}{\frac{1}{G}\sum^G_{i=1}\left|r_{i}-\operatorname{mean}\left(\left\{r_{i}\right\}^G_{i=1}\right)\right|}\right| = G.
\end{equation}
\end{theorem}

Crucially, Theorem~\ref{th:dgae} removes the binary reward constraint ($r_i \in \{0,1\}$) in Theorem~\ref{th:grae}, rendering it suitable for a wide array of policy optimization scenarios.

\vspace{-0.05in}
\subsubsection{Difficulty-Aware Question-Level Weighting}
\vspace{-0.05in}
% DQW

Building upon the DGAE strategy, we further introduce a difficulty-aware question-level weighting (DQW) scheme, which explicitly prioritizes learning from more challenging questions within each training batch.
Specifically, DQW assigns a weight $\lambda_s$ to each question $q_s$ as follows:
\begin{equation}
    \lambda_s = B_\text{v}\cdot\frac{\exp\left(D_s/T\right)}{\sum^{B_\text{v}}_{s=1}\exp\left(D_s/T\right)},~~\text{where}~D_s=-\operatorname{mean}\left(\left\{r_{si}\right\}^G_{i=1}\right).
\end{equation}
Here, $D_s$ is the negative mean reward across all responses of the question $q_s$, serving as a measure of its relative difficulty at the current training stage, and $T$ denotes the temperature hyperparameter that controls the distribution sharpness.
Compared to advantage reweighting of \citet{zhang2025grpo}, DQW is simpler and has fewer hyperparameters.
Moreover, it is derived based on the analysis of the implicit update magnitude imbalance in GRPO and the balanced advantages of DGAE.
This two-step “balance-then-reweight” procedure offers improved interpretability and controllability.

\vspace{-0.05in}
\subsection{Multi-Aspect Question Reformulation}
\vspace{-0.05in}

DGPO enhances mathematical reasoning from an algorithmic perspective by optimizing the learning process on existing data.
To complement this, we propose the Multi-Aspect Question Reformulation (MQR) approach as a data-centric solution, which automatically reformulates training questions by a large reasoning model to generate variants that cover more complex and comprehensive aspects.
A critical constraint is that \textit{all reformulations must preserve the original gold answer}.
In this manner, MQR can maintain the essential mathematical logic of the question and obviate the need for solution regeneration, thereby placing minimal demands on the reformulator model.

Specifically, MQR adds story background, introduces abstract terminology, and nests sub-problems into the original question.
The default reformulator model is OpenAI o3, while some smaller open-source models can also competently handle this task.
The prompts are provided in Appendix~\ref{sec:appendix_prompts}, and the core instructions for these strategies are as follows:
\begin{promptbox}{Core Instructions for Multi-Aspect Question Reformulation}
\begin{enumerate}[leftmargin=12pt, topsep=-4pt, itemsep=0pt, partopsep=0pt]
    \item \textbf{Background:} Add a story background that is not related to the core mathematical content of the given question, but seems to be related to the question. If the given question already has such a background, change it to a new, complexer background.
    \item \textbf{Term:} Invent a new, abstract mathematical term to define a concept that is central to the given question, and restate the entire question using this term.
    \item \textbf{Sub-Problem:} Convert a key numerical condition of the given question which have a definite value into an independent sub-problem. The sub-problem may belong to any branch of mathematics (e.g., algebra, geometry, number theory, combinatorics).
\end{enumerate}
\end{promptbox}
The newly generated questions respectively challenge the policy model's ability to: 1) identify critical mathematical information amidst noise; 2) grasp abstract mathematical concepts; and 3) perform reasoning that requires multiple steps and cross-domain knowledge.
Successfully solving these more difficult questions provides a strong reinforcement signal, compelling the policy model to develop these crucial reasoning skills.
Examples of each reformulation aspect are provided in Appendix~\ref{sec:appendix_data}.

Overall, the MQR-augmented data, which combines the original and reformulated questions, serves as ideal training material for DGPO, rendering MathForge a synergistic loop where the data extends the model's performance boundaries, and the algorithm efficiently learns from these challenges.

\vspace{-0.05in}
\section{Experiments}
\vspace{-0.05in}

\vspace{-0.05in}
\subsection{Experimental Setup}\label{sec:setup}
\vspace{-0.05in}

\textbf{Models and Datasets.}~~In the main experiments, we train the Qwen2.5-Math-7B model \citep{yang2024qwen2} on the MATH dataset \citep{hendrycks2021measuring}.
To evaluate the model-agnostic effectiveness of MathForge, we conduct experiments on three other models of varying sizes and types: Qwen2.5-Math-1.5B \citep{yang2024qwen2}, Qwen2.5-3B \citep{qwen2025qwen2}, and DeepSeek-Math-7B \citep{shao2024deepseekmath}.
For cold start, DeepSeek-Math-7B is fine-tuned using 80k data sampled from NuminaMath-CoT \citep{li2024numinamath}.
Furthermore, we apply DGPO in the multimodal domain, training Qwen2.5-VL-3B-Instruct \citep{bai2025qwen2} on the GEOQA-8k dataset \citep{chen2025r1v}.

\textbf{Benchmarks.}~~In the text-only experiments, we assess models on six commonly used mathematical reasoning benchmarks: AIME24, AIME25, AMC23, MATH500 \citep{hendrycks2021measuring}, Minerva \citep{lewkowycz2022solving}, and Olympiad \citep{he2024olympiadbench}.
To ensure stable results, we perform 32 runs for AIME24, AIME25, and AMC-23, and 4 runs for other benchmarks, reporting the average performance across the respective runs.
For the multimodal domain, we evaluate on the GeoQA test set \citep{chen2021geoqa} using greedy decoding.
All evaluations are conducted in a zero-shot setting.

\textbf{Compared Methods.}~~We compare our MathForge framework against several state-of-the-art methods: GRPO \citep{shao2024deepseekmath}, Dr.GRPO \citep{liu2025understanding}, GPG \citep{chu2025gpg}, DAPO \citep{yu2025dapo}, GSPO \citep{zheng2025group}, and GRPO-AD \citep{zhang2025grpo}.
For a fair algorithm-level comparison, we disable the resampling components in GPG and DAPO, and add the Advantage reweighting for Difficulty (AD) technique of \citet{zhang2025grpo} into the GRPO baseline as GRPO-AD.
To isolate the contribution of each component in MathForge, we also evaluate DGPO and MQR separately.
The MQR setting refers to training on the MQR-augmented data, including the original and MQR-generated data, using GRPO.

\begin{table}[t!]
    \vspace{-0.05in}
    \caption{Comparative results of methods trained on the MATH dataset using Qwen2.5-Math-7B.}
    \label{tab:performance}
    \vspace{-0.03in}
    \begin{center}
    \scalebox{0.88}{
    \tabcolsep7.3pt
    {\renewcommand{\arraystretch}{1.2}
        \begin{tabular}{l|cccccc|l}
        \toprule[1.2pt]
        Methods & AIME24 & AIME25 & AMC23 & MATH500 & Minerva & Olympiad & Avg./$\Delta_\text{GRPO}$ \\
        \midrule
        Base Model & 12.19 & 4.79 & 35.23 & 48.60 & 15.07 & 16.33 & ~~22.04 \\
        \midrule
        GRPO & 20.94 & 8.44 & 58.98 & 72.20 & 27.76 & 37.33 & ~~37.61 \\
        Dr.GRPO & 21.04 & 8.23 & 58.59 & 72.05 & 28.58 & 35.89 & ~~37.40\tiny{$\textcolor{down}{-0.21}$} \\
        GPG & 21.98 & 9.06 & 59.61 & 72.05 & 27.21 & 37.67 & ~~37.93\tiny{$\textcolor{up}{+0.32}$} \\
        DAPO & 21.25 & 8.75 & 58.20 & 72.70 & 29.50 & 37.22 & ~~37.94\tiny{$\textcolor{up}{+0.33}$} \\
        GSPO & 19.38 & 8.33 & \underline{60.16} & 73.00 & 28.12 & 37.26 & ~~37.71\tiny{$\textcolor{up}{+0.10}$} \\
        GRPO-AD & 21.56 & 9.48 & 59.06 & 73.25 & 29.14 & 37.07 & ~~38.26\tiny{$\textcolor{up}{+0.65}$} \\
        \rowcolor{lightblue} DGPO & 23.85 & 10.21 & \textbf{61.02} & 74.25 & 31.07 & 38.33 & ~~39.79\tiny{$\textcolor{up}{+2.18}$} \\
        \rowcolor{lightblue} MQR & \textbf{25.00} & \underline{11.77} & 59.38 & \underline{77.85} & \underline{31.43} & \underline{40.81} & ~~\underline{41.04}\tiny{$\textcolor{up}{+3.43}$} \\
        \rowcolor{lightblue} MathForge & \underline{24.58} & \textbf{12.60} & 59.84 & \textbf{79.95} & \textbf{33.36} & \textbf{42.67} & ~~\textbf{42.17}\tiny{$\textcolor{up}{+4.56}$} \\
        \bottomrule[1.2pt]
        \end{tabular}}}
    \end{center}
    \vspace{-0.2in}
\end{table}

\textbf{Implementation Details.}~~We used 8 NVIDIA H20 GPUs to conduct all experiments.
To ensure fair comparison and reproducibility, our implementation is built upon the Open-R1 codebase \citep{openr1}.
For the DGPO algorithm, the temperature hyperparameter $T$ in the DQW component is set to $2.0$.
For the MQR strategy, the data augmentation cost is reported in Appendix~\ref{sec:appendix_cost}.
All other implementation details are provided in Appendix~\ref{sec:appendix_implementation}.

\vspace{-0.05in}
\subsection{Main Results}
\vspace{-0.05in}

Table~\ref{tab:performance} presents the comparative results of various methods trained on the MATH dataset using the Qwen2.5-Math-7B model. In the following, we will analyze the effectiveness of DGPO, MQR, and their combination, MathForge, respectively.

\begin{table}[t!]
    \vspace{-0.05in}
    \caption{Comparative results of methods trained on the MATH dataset using varying base models.}
    \label{tab:models}
    \vspace{-0.01in}
    \begin{center}
    \scalebox{0.88}{
    \tabcolsep5.1pt
    {\renewcommand{\arraystretch}{1.2}
        \begin{tabular}{l|cccccc|l}
        \toprule[1.2pt]
        Methods & AIME24 & AIME25 & AMC23 & MATH500 & Minerva & Olympiad & Avg./$\Delta_\text{GRPO}$ \\
        \midrule
        Qwen2.5-Math-1.5B & 6.87 & 3.65 & 30.94 & 34.95 & 8.55 & 21.93 & ~~17.82 \\
        ~+ GRPO & 11.35 & 3.96 & 46.48 & 64.85 & 20.13 & 29.59 & ~~29.39 \\
        \rowcolor{lightblue}~+ DGPO & 11.25 & \underline{5.73} & 49.84 & 65.45 & 21.14 & 30.85 & ~~30.71\tiny{$\textcolor{up}{+1.32}$} \\
        \rowcolor{lightblue}~+ MQR & \underline{11.98} & 5.42 & \underline{50.08} & \underline{69.65} & \underline{23.81} & \underline{33.67} & ~~\underline{32.44}\tiny{$\textcolor{up}{+3.05}$} \\
        \rowcolor{lightblue}~+ MathForge & \textbf{13.23} & \textbf{7.71} & \textbf{52.34} & \textbf{70.10} & \textbf{25.74} & \textbf{33.89} & ~~\textbf{33.84}\tiny{$\textcolor{up}{+4.45}$} \\
        \midrule
        Qwen2.5-3B & 2.81 & 0.73 & 22.66 & 48.65 & 13.69 & 19.37 & ~~17.99 \\
        ~+ GRPO & 5.31 & \underline{1.56} & 33.28 & 63.35 & 22.89 & 26.41 & ~~25.47 \\
        \rowcolor{lightblue}~+ DGPO & \textbf{6.98} & \underline{1.56} & 36.56 & \textbf{65.80} & 25.28 & 26.96 & ~~27.19\tiny{$\textcolor{up}{+1.72}$} \\
        \rowcolor{lightblue}~+ MQR & 5.10 & \underline{1.56} & \underline{39.53} & 65.20 & \underline{25.74} & \underline{29.19} & ~~\underline{27.72}\tiny{$\textcolor{up}{+2.25}$} \\
        \rowcolor{lightblue}~+ MathForge & \underline{5.73} & \textbf{1.77} & \textbf{40.70} & \underline{65.40} & \textbf{28.86} & \textbf{31.59} & ~~\textbf{29.01}\tiny{$\textcolor{up}{+3.54}$} \\
        \midrule
        DeepSeek-Math-7B & 0.42 & 0.10 & 13.28 & 31.05 & 9.56 & 9.00 & ~~10.57 \\
        ~+ GRPO & 0.63 & 0.10 & 19.14 & 41.45 & 14.71 & 13.44 & ~~14.91 \\
        \rowcolor{lightblue}~+ DGPO & \underline{1.98} & 0.42 & \underline{21.02} & 41.85 & \underline{18.93} & 15.00 & ~~16.53\tiny{$\textcolor{up}{+1.62}$} \\
        \rowcolor{lightblue}~+ MQR & \underline{1.98} & \textbf{0.83} & 20.86 & \textbf{44.25} & 17.00 & \underline{15.74} & ~~\underline{16.78}\tiny{$\textcolor{up}{+1.87}$} \\
        \rowcolor{lightblue}~+ MathForge & \textbf{3.12} & \underline{0.73} & \textbf{21.72} & \underline{43.60} & \textbf{20.68} & \textbf{16.74} & ~~\textbf{17.77}\tiny{$\textcolor{up}{+2.86}$} \\
        \bottomrule[1.2pt]
        \end{tabular}}}
    \end{center}
    \vspace{-0.2in}
\end{table}

\textbf{Effectiveness of DGPO.}~~Our DGPO algorithm, when applied alone, elevates the average score to 39.79\%, a substantial gain of 2.18\% over the strong GRPO baseline (37.61\%). 
This result validates our hypothesis that prioritizing more challenging questions through DGAE and DQW effectively enhances the RL training process. 
By rectifying the update magnitude imbalance of GRPO and explicitly focusing the model on its solvable weakness, DGPO fosters a more efficient and targeted optimization.
Additionally, DGPO also surpasses other advanced policy optimization techniques, highlighting the superior design and efficacy of our proposed difficulty-aware mechanisms.

\begin{table}[t!]
    \vspace{-0.06in}
    \caption{Ablation Results of DGPO trained on the MATH dataset using Qwen2.5-Math-7B.}
    \label{tab:ablation}
    \vspace{-0.01in}
    \begin{center}
    \scalebox{0.88}{
    \tabcolsep2.9pt
    {\renewcommand{\arraystretch}{1.2}
        \begin{tabular}{l|cccccc|l}
        \toprule[1.2pt]
        Methods & AIME24 & AIME25 & AMC23 & MATH500 & Minerva & Olympiad & Avg./$\Delta_\text{GRPO}$ \\
        \midrule
        GRPO & 20.94 & 8.44 & 58.98 & 72.20 & 27.76 & 37.33 & ~~37.61 \\
        DGPO (w/o DGAE \& DQW) & 20.21 & 9.06 & 59.45 & 72.40 & 28.58 & 36.56 & ~~37.71\tiny{$\textcolor{up}{+0.10}$} \\
        DGPO (w/o DQW) & \underline{21.77} & \underline{9.69} & \underline{60.00} & \underline{73.45} & \underline{29.04} & \underline{37.93} & ~~\underline{38.65}\tiny{$\textcolor{up}{+1.04}$} \\
        \rowcolor{lightblue} DGPO (full) & \textbf{23.85} & \textbf{10.21} & \textbf{61.02} & \textbf{74.25} & \textbf{31.07} & \textbf{38.33} & ~~\textbf{39.79}\tiny{$\textcolor{up}{+2.18}$} \\
        \midrule
        DGPO ($T=1.0$) & \underline{23.12} & 9.06 & 59.45 & 74.15 & \underline{30.61} & 37.78 & ~~39.03\tiny{$\textcolor{up}{+1.42}$} \\
        \rowcolor{lightblue} DGPO ($T=2.0$) & \textbf{23.85} & \underline{10.21} & \underline{61.02} & \underline{74.25} & \textbf{31.07} & \textbf{38.33} & ~~\textbf{39.79}\tiny{$\textcolor{up}{+2.18}$}\\
        DGPO ($T=5.0$) & 22.81 & \textbf{11.35} & 60.55 & 73.80 & 30.42 & \underline{38.26} & ~~\underline{39.53}\tiny{$\textcolor{up}{+1.92}$} \\
        DGPO ($T=10.0$) & 21.35 & 9.79 & \textbf{62.27} & \textbf{74.55} & 29.96 & 37.67 & ~~39.27\tiny{$\textcolor{up}{+1.66}$} \\
        \bottomrule[1.2pt]
        \end{tabular}}}
    \end{center}
    \vspace{-0.2in}
\end{table}

\textbf{Effectiveness of MQR.}~~The use of MQR in training also yields significant improvements, reaching an average score of 41.04\%, which is a 3.43\% increase over GRPO. 
This demonstrates the validity of our three question reformulation strategies. 
By augmenting the training data with questions that introduce narrative noise (Background), abstract concepts (Term), and nested logic (Sub-Problem), MQR creates a more challenging and diverse learning environment. 
This substantial performance improvement indicates the effectiveness of compelling the model to develop more robust reasoning skills by tackling these more complex reformulated questions.

\textbf{Effectiveness of MathForge.}~~The combination of DGPO and MQR in the full MathForge framework achieves the best overall performance, outperforming both individual components and reaching an average of 42.17\%.
This result highlights a powerful synergy between the data-centric and algorithmic components of our framework. 
MQR provides the ideal training material, diverse and challenging questions that expose the model's limitations, while DGPO capitalizes on this data by ensuring the model focuses its updates on mastering these challenges.
Additionally, the performance gaps between DGPO and GRPO, as well as between MathForge and MQR, further demonstrate the robustness of DGPO under different query difficulty, as MQR makes questions harder.

\textbf{Model-Agnosticism of MathForge.}~~To substantiate the model-agnosticism of MathForge, we further compare methods across different model sizes and types, as presented in Table~\ref{tab:models}.
MathForge consistently delivers the best performance on all models, and the individual components, DGPO and MQR, also robustly outperform GRPO.
This highlights that the principles of MathForge are fundamental and not contingent on a specific model, underscoring its broad generalizability.

\vspace{-0.05in}
\subsection{Analysis of DGPO}
\vspace{-0.05in}

\textbf{Ablation Studies.}~~As shown in Table~\ref{tab:ablation}, we conduct ablation experiments to isolate the contribution of each component in DGPO.
Specifically, the valid token-level loss averaging, DGAE, and DQW components contribute average performance improvements of 0.10\%, 0.94\%, and 1.14\%, respectively.
This highlights that DGAE effectively corrects the update magnitude imbalance of GRPO, and DQW provides a significant and complementary benefit by explicitly prioritizing more challenging questions.
Additionally, we investigate the sensitivity of the temperature hyperparameter $T$ in DQW.
The results indicate that $T=2.0$ yields the best overall performance.
A lower temperature may potentially lead to an overly sharp distribution that focuses too narrowly on the hardest question in a batch, while a higher temperature flattens the weighting distribution, diminishing the prioritization effect of DQW.
This confirms that $T=2.0$ strikes an optimal balance, effectively emphasizing difficult questions while maintaining sufficient learning from the entire batch.
Because the difficulty score is bounded within $(-1, 0)$, setting $T=2.0$ ensures that the ratio between the maximum and minimum weights in a batch remains below $e^{0/T}/e^{-1/T} = e^{1/2} \approx 1.65$.

\begin{wrapfigure}{r}{0.6\textwidth}
    \vspace{-0.2in}
    \centering
    \subfigure[Accuracy Reward.]{
        \includegraphics[width=0.4725\linewidth]{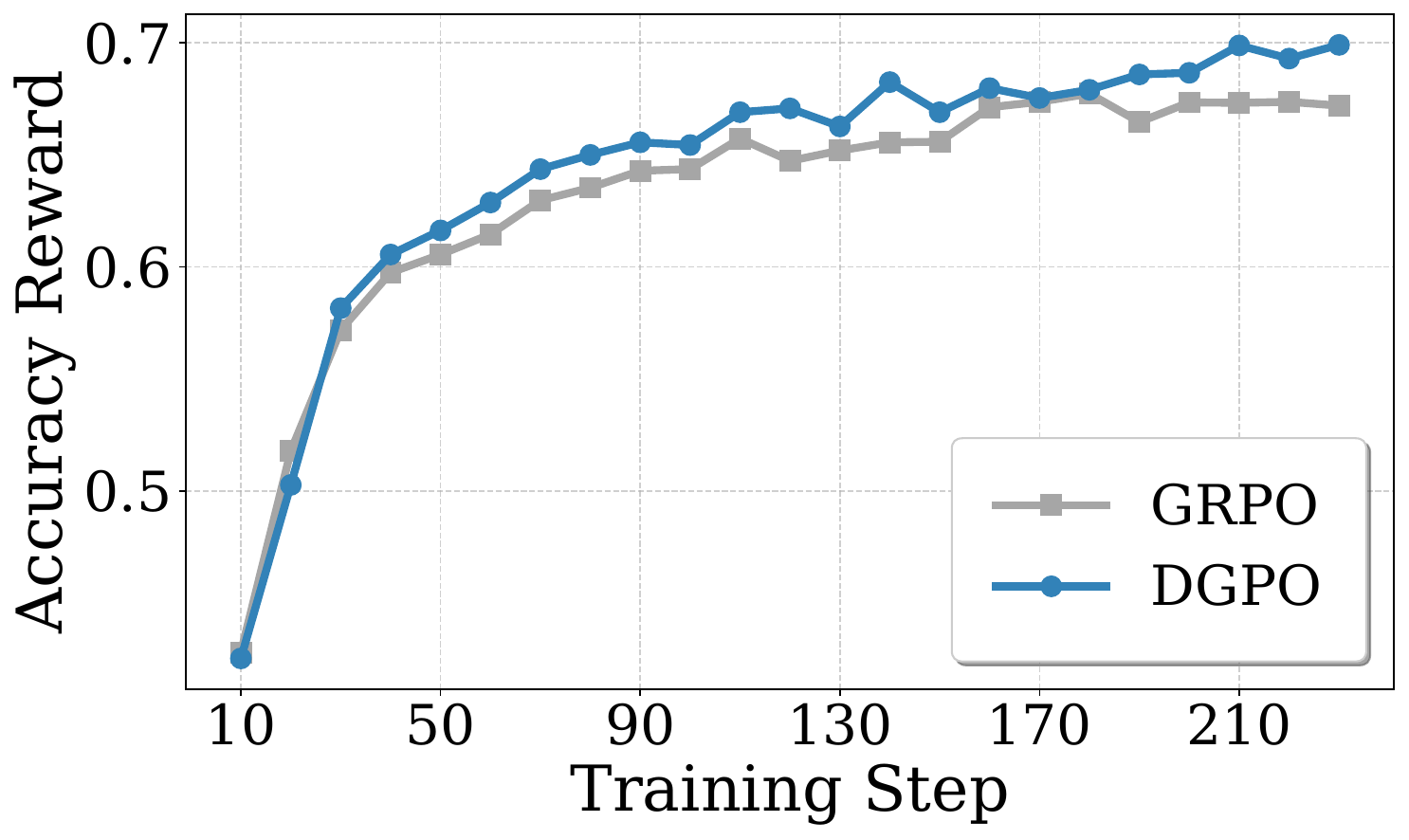}
        \label{fig:accuracy}
    }
    \subfigure[Output Length.]{
        \includegraphics[width=0.4675\linewidth]{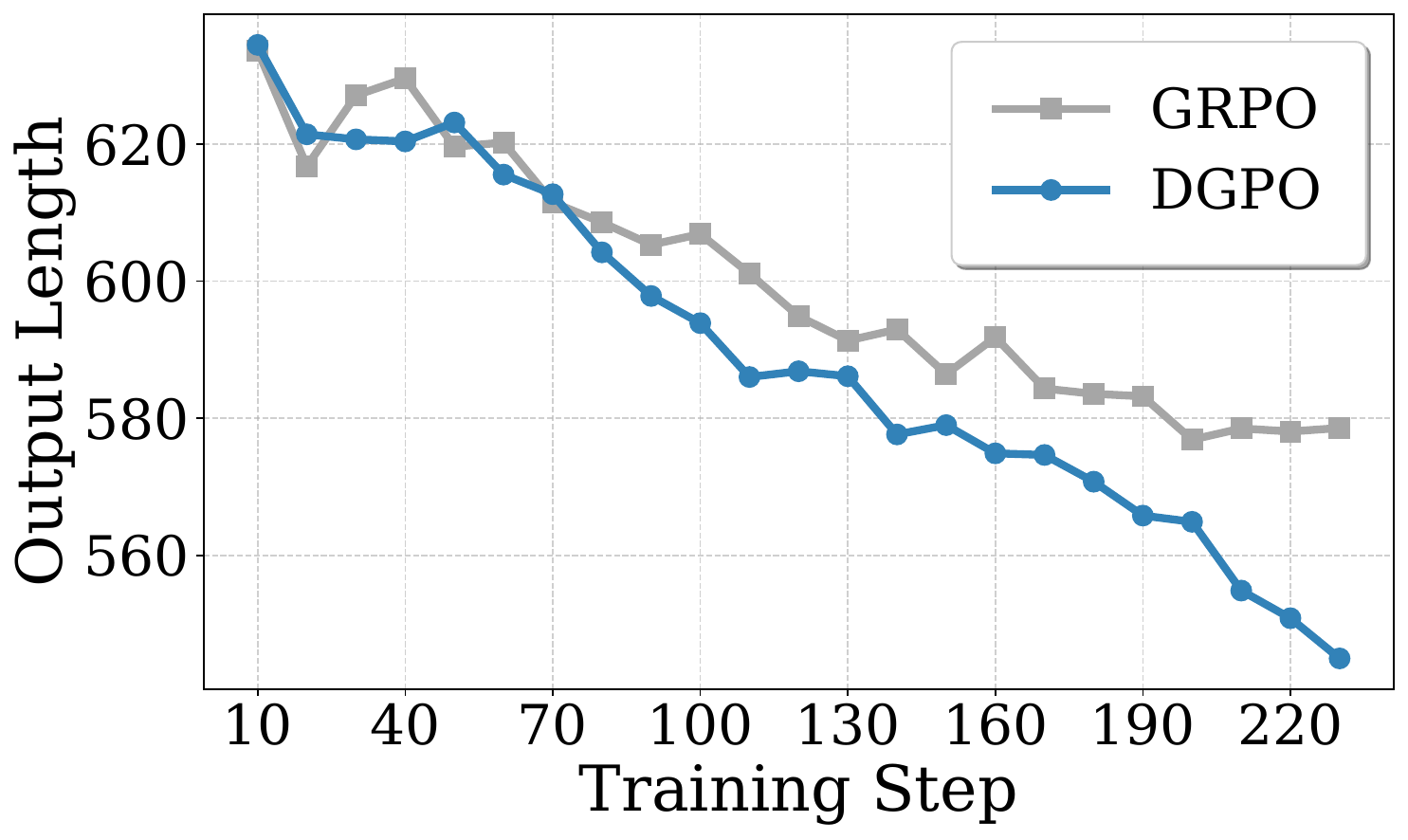}
        \label{fig:length}
    }
    \vspace{-0.2in}
    \caption{Training dynamics of DGPO vs. GRPO evaluated on the MATH500 benchmark. Both models are trained on MATH using Qwen2.5-Math-7B.}
    \label{fig:dynamic}
    \vspace{-0.1in}
\end{wrapfigure}

\textbf{Training Dynamics.}~~Figure~\ref{fig:dynamic} shows the training dynamics of DGPO versus GRPO in our main experiments, illustrating the evolution of accuracy rewards and model output lengths on MATH500.
As demonstrated in Figure~\ref{fig:accuracy}, DGPO consistently outperforms GRPO after the initial phase, and the performance gap widens as training progresses, underscoring that prioritizing harder questions leads to a more substantial and sustained improvement in accuracy.
Meanwhile, Figure~\ref{fig:length} indicates that DGPO tends to produce more concise responses, highlighting that DGPO not only improves correctness but also encourages the model to find more efficient and direct reasoning paths, trimming unnecessary verbosity and redundant steps.

\begin{table}[t!]
    \vspace{-0.05in}
    \caption{Synergistic results of DGPO with other policy optimization methods trained on the MATH dataset using Qwen2.5-Math-7B.}
    \label{tab:combination}
    \vspace{-0.02in}
    \begin{center}
    \scalebox{0.88}{
    \tabcolsep9pt
    {\renewcommand{\arraystretch}{1.2}
        \begin{tabular}{l|cccccc|c}
        \toprule[1.2pt]
        Methods & AIME24 & AIME25 & AMC23 & MATH500 & Minerva & Olympiad & Average \\
        \midrule
        GPG & \textbf{21.98} & 9.06 & 59.61 & 72.05 & 27.21 & 37.67 & 37.93 \\
        \rowcolor{lightblue}~+ DGPO & 21.77 & \textbf{10.00} & \textbf{60.00} & \textbf{73.45} & \textbf{30.06} & \textbf{38.26} & \textbf{38.92} \\
        \midrule
        DAPO & 21.25 & 8.75 & 58.20 & 72.70 & 29.50 & 37.22 & 37.94 \\
        \rowcolor{lightblue}~+ DGPO & \textbf{24.48} & \textbf{9.79} & \textbf{58.75} & \textbf{74.90} & \textbf{31.99} & \textbf{39.56} & \textbf{39.91} \\
        \midrule
        GSPO & 19.38 & 8.33 & \textbf{60.16} & 73.00 & 28.12 & 37.26 & 37.71 \\
        \rowcolor{lightblue}~+ DGPO & \textbf{23.33} & \textbf{10.00} & 59.14 & \textbf{74.15} & \textbf{30.88} & \textbf{38.41} & \textbf{39.32} \\
        \bottomrule[1.2pt]
        \end{tabular}}}
    \end{center}
    \vspace{-0.2in}
\end{table}

\begin{table}[t!]
    \vspace{-0.06in}
    \caption{Comparative results of methods trained on the GEOQA-8k dataset using Qwen2.5-VL-3B-Instruct in the multimodal domain.}
    \label{tab:vl}
    % \vspace{-0.02in}
    \begin{center}
    \scalebox{0.88}{
    \tabcolsep3pt
    {\renewcommand{\arraystretch}{1.2}
        \begin{tabular}{l|c|ccccccc}
        \toprule[1.2pt]
        Methods & Base Model & GRPO & Dr.GRPO & GPG & DAPO & GSPO & GRPO-AD & \cellcolor{lightblue}DGPO \\
        \midrule
        GeoQA/$\Delta_\text{GRPO}$ & 39.79 & 57.43 & 57.96\tiny{$\textcolor{up}{+0.53}$} & \underline{59.02}\tiny{$\textcolor{up}{+1.59}$} & \underline{59.02}\tiny{$\textcolor{up}{+1.59}$} & 57.16\tiny{$\textcolor{down}{-0.27}$} & 58.09\tiny{$\textcolor{up}{+0.66}$} & \cellcolor{lightblue}\textbf{59.95}\tiny{$\textcolor{up}{+2.52}$} \\
        \bottomrule[1.2pt]
        \end{tabular}}}
    \end{center}
    \vspace{-0.2in}
\end{table}

\textbf{Compatibility with Other Methods.}~~Our DGPO algorithm primarily introduces an improved advantage estimation and an additional question-level weighting scheme, both of which are compatible with most existing policy optimization methods.
To demonstrate this, we integrate DGPO with GPG, DAPO, and GSPO, respectively.
The combination forms are detailed in Appendix~\ref{sec:appendix_combination}.
As shown in Table~\ref{tab:combination}, this integration yields consistent performance improvements.
Particularly, the combination of DAPO with DGPO results in even higher performance than the standalone DGPO (39.91\% vs. 39.79\%).
This underscores that DGPO addresses a fundamental aspect of the learning process that complements the specific mechanics of other policy optimization methods. 
In other words, DGPO can function as a general enhancement algorithm rather than a monolithic alternative.

\textbf{Applicability in the Multimodal Domain.}~~To further verify the domain-agnosticism of DGPO, we apply it to a multimodal mathematical reasoning task. 
As shown in Table~\ref{tab:vl}, DGPO achieves the best performance of 59.95\% again, significantly higher than that of GRPO (57.43\%).
This demonstrates that the core principle of our DGPO, prioritizing more challenging questions, is not confined to text-only reasoning.
It is a robust and generalizable algorithm for enhancing policy learning wherever a quantifiable measure of difficulty (such as accuracy rate) can be established.

\vspace{-0.05in}
\subsection{Analysis of MQR}
\vspace{-0.05in}

In this subsection, we normalize the total training data volume across all methods for a fair comparison. 
Since MQR expands the dataset by a factor of four, we achieve this by increasing the training epochs for each method accordingly.
As shown in Table~\ref{tab:models_mqr}, we compare the performance of methods trained on the original data versus the MQR-augmented data using DGPO and varying base models.
MQR consistently yields superior results than the original data across all models, confirming that its effectiveness stems from the qualitative enhancement of the data, not merely an increase in volume.
Additionally, we assess the quality of the generated data in Appendix~\ref{sec:appendix_quality}.

\textbf{Difficulty Assessment.}~~We first conduct a direct comparison of question difficulty by evaluating the accuracy of Qwen2.5-Math-7B-Instruct on the subsets of MQR-augmented data.
The accuracy rates are 79.77\% on Original, 77.31\% on Background, 76.87\% on Term, and 72.04\% on Sub-Problem, confirming the increased difficulty of reformulated questions and the effectiveness of MQR.

\textbf{Ablation Studies.}~~To assess the individual contributions of our three reformulation strategies, we conduct ablation studies where each strategy is utilized separately.
MetaMath-Rephrasing \citep{yu2024metamath} is also included as a baseline, which uses GPT-3.5-Turbo to simply rephrase questions. We sample 22.5k data from its total 50k rephrased questions, combined with the original data for training.
The results, as presented in Table~\ref{tab:performance_mqr}, are all trained using DGPO.
Each strategy independently improves performance over both the Original and the MetaMath-Rephrasing baselines.
Crucially, the MQR approach, which combines all three strategies, achieves the highest average score of 42.17\%. 
This underscores a clear synergy, where these diverse strategies produce a more substantial improvement than any individual component in mathematical reasoning.

\begin{table}[t!]
    \vspace{-0.05in}
    \caption{Comparative results of methods trained on the original data vs. the MQR-augmented data using DGPO and varying base models.}
    \label{tab:models_mqr}
    \vspace{-0.03in}
    \begin{center}
    \scalebox{0.88}{
    \tabcolsep4.2pt
    {\renewcommand{\arraystretch}{1.2}
        \begin{tabular}{l|l|cccccc|c}
        \toprule[1.2pt]
        Models & Data & AIME24 & AIME25 & AMC23 & MATH500 & Minerva & Olympiad & Average \\
        \midrule
        \multirow{2}{*}{Qwen2.5-Math-7B} & Ori. & \textbf{26.46} & 9.17 & 58.67 & 74.65 & 31.62 & 38.81 & 39.90  \\
        \multirow{2}{*}{} & \cellcolor{lightblue}MQR & \cellcolor{lightblue}24.58 & \cellcolor{lightblue}\textbf{12.60} & \cellcolor{lightblue}\textbf{59.84} & \cellcolor{lightblue}\textbf{79.95} & \cellcolor{lightblue}\textbf{33.36} & \cellcolor{lightblue}\textbf{42.67} & \cellcolor{lightblue}\textbf{42.17} \\
        \midrule
        \multirow{2}{*}{Qwen2.5-Math-1.5B} & Ori. & 11.98 & 5.21 & 50.62 & 68.40 & 24.26 & 32.59 & 32.18 \\
        \multirow{2}{*}{} & \cellcolor{lightblue}MQR & \cellcolor{lightblue}\textbf{13.23} & \cellcolor{lightblue}\textbf{7.71} & \cellcolor{lightblue}\textbf{52.34} & \cellcolor{lightblue}\textbf{70.10} & \cellcolor{lightblue}\textbf{25.74} & \cellcolor{lightblue}\textbf{33.89} & \cellcolor{lightblue}\textbf{33.84} \\
        \midrule
        \multirow{2}{*}{Qwen2.5-3B} & Ori. & \textbf{6.04} & 1.35 & 37.66 & 65.05 & 25.28 & 27.93 & 27.22 \\
        \multirow{2}{*}{} & \cellcolor{lightblue}MQR & \cellcolor{lightblue}5.73 & \cellcolor{lightblue}\textbf{1.77} & \cellcolor{lightblue}\textbf{40.70} & \cellcolor{lightblue}\textbf{65.40} & \cellcolor{lightblue}\textbf{28.86} & \cellcolor{lightblue}\textbf{31.59} & \cellcolor{lightblue}\textbf{29.01} \\
        \midrule
        \multirow{2}{*}{DeepSeek-Math-7B} & Ori. & 2.19 & 0.21 & 21.02 & \textbf{43.60} & 18.29 & 14.52 & 16.64 \\
        \multirow{2}{*}{} & \cellcolor{lightblue}MQR & \cellcolor{lightblue}\textbf{3.12} & \cellcolor{lightblue}\textbf{0.73} & \cellcolor{lightblue}\textbf{21.72} & \cellcolor{lightblue}\textbf{43.60} & \cellcolor{lightblue}\textbf{20.68} & \cellcolor{lightblue}\textbf{16.74} & \cellcolor{lightblue}\textbf{17.77} \\
        \bottomrule[1.2pt]
        \end{tabular}}}
    \end{center}
    \vspace{-0.2in}
\end{table}

\begin{table}[t!]
    \vspace{-0.1in}
    \caption{Ablation Results of MQR on the MATH dataset using Qwen2.5-Math-7B.}
    \label{tab:performance_mqr}
    \vspace{-0.03in}
    \begin{center}
    \scalebox{0.88}{
    \tabcolsep4.8pt
    {\renewcommand{\arraystretch}{1.2}
        \begin{tabular}{l|cccccc|l}
        \toprule[1.2pt]
        Data & AIME24 & AIME25 & AMC23 & MATH500 & Minerva & Olympiad & Avg./$\Delta_\text{Ori.}$ \\
        \midrule
        Original & \underline{26.46} & 9.17 & 58.67 & 74.65 & 31.62 & 38.81 & ~~39.90  \\
        MetaMath-Rephrasing & 25.21 & \underline{11.35} & \underline{59.45} & 76.70 & 31.71 & 39.93 & ~~40.73\tiny{$\textcolor{up}{+0.83}$} \\
        Original + Background & 25.52 & 10.73 & 58.59 & 77.50 & 32.90 & 40.48 & ~~40.95\tiny{$\textcolor{up}{+1.05}$}  \\
        Original + Term & 25.52 & 11.15 & 58.98 & \underline{77.75} & 33.09 & 40.93 & ~~41.24\tiny{$\textcolor{up}{+1.34}$}  \\
        Original + Sub-Problem & \textbf{26.67} & 10.94 & 58.75 & 77.05 & \textbf{34.38} & \underline{41.36} & ~~\underline{41.53}\tiny{$\textcolor{up}{+1.63}$} \\
        \rowcolor{lightblue} MQR & 24.58 & \textbf{12.60} & \textbf{59.84} & \textbf{79.95} & \underline{33.36} & \textbf{42.67} & ~~\textbf{42.17}\tiny{$\textcolor{up}{+2.27}$} \\
        \bottomrule[1.2pt]
        \end{tabular}}}
    \end{center}
    \vspace{-0.21in}
\end{table}

\begin{wrapfigure}{r}{0.6\textwidth}
    \vspace{-0.2in}
    \centering
    \subfigure[Training.]{
        \includegraphics[width=0.47\linewidth]{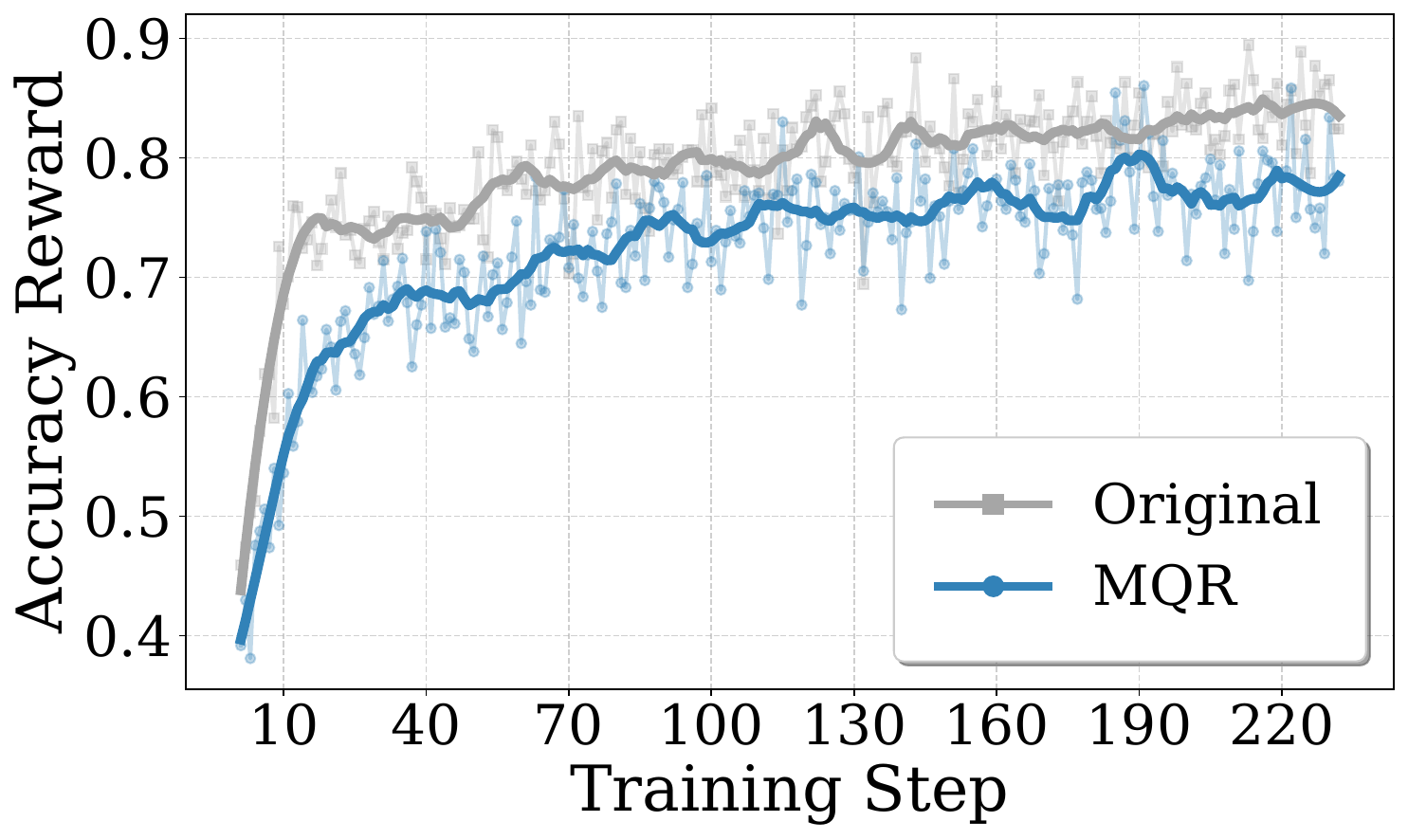}
        \label{fig:mqr_train}
    }
    \subfigure[Evaluation.]{
        \includegraphics[width=0.47\linewidth]{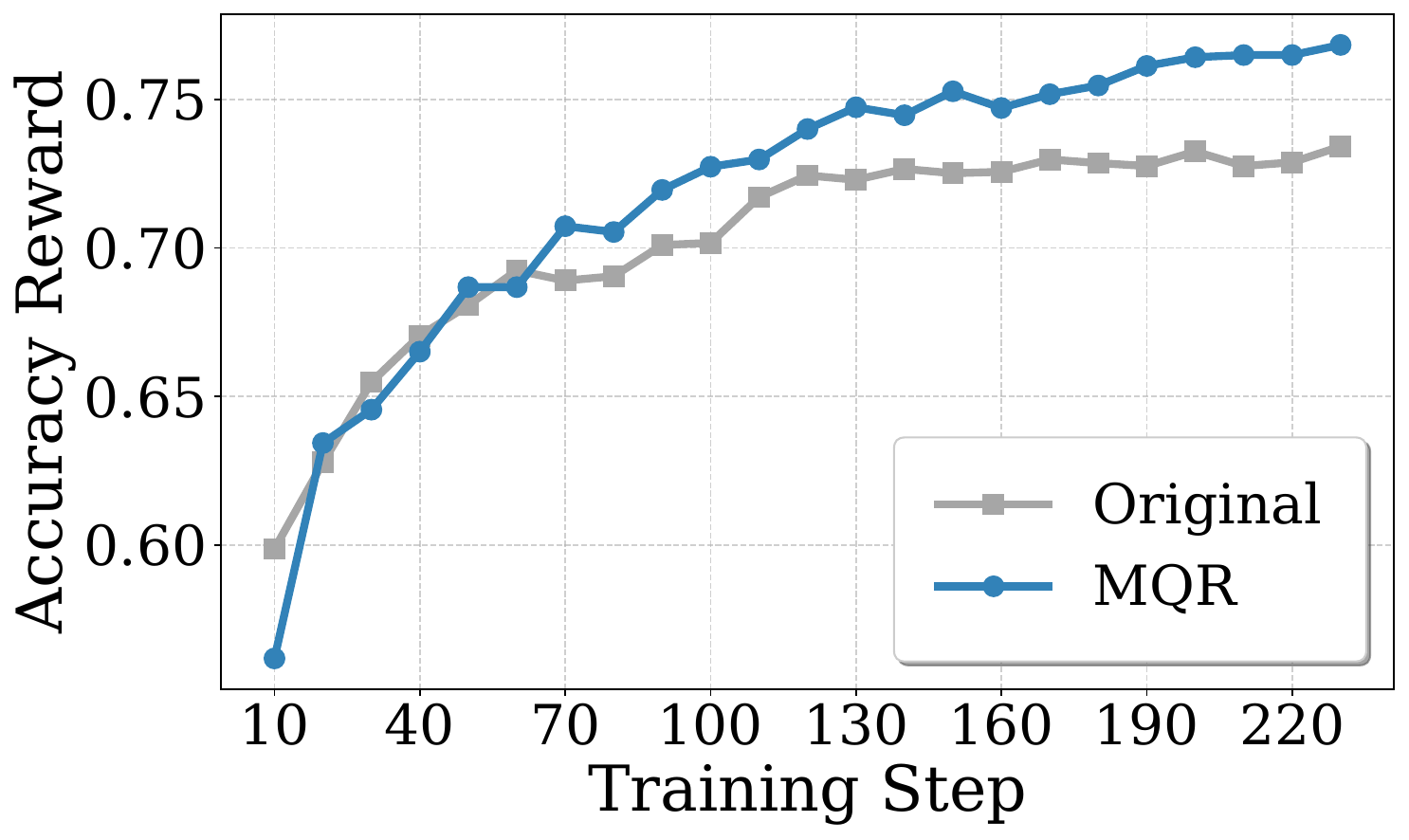}
        \label{fig:mqr}
    }
    \vspace{-0.2in}
    \caption{Training dynamics of Original vs. MQR on training and evaluation data. Both models are trained on MATH and evaluated on MATH500 using Qwen2.5-Math-7B.}
    \label{fig:dynamic_mqr}
    \vspace{-0.1in}
\end{wrapfigure}

\textbf{Training Dynamics.}~~Figure~\ref{fig:dynamic_mqr} illustrates the training dynamics of DGPO when trained on the original MATH dataset versus the MQR-augmented dataset.
As presented in Figure~\ref{fig:mqr_train}, the consistently lower training accuracy on the MQR-augmented data exhibits that the reformulated questions are substantially more challenging.
Despite this increased difficulty, the model trained with MQR ultimately achieves superior accuracy on the unseen MATH500 benchmark, as depicted in Figure~\ref{fig:mqr}.
This ``train harder, test better" phenomenon suggests that the more challenging questions of MQR result in robust, generalizable reasoning capabilities, enhancing performance while preventing overfitting.

\textbf{Generality to Less Capable Reformulators.}~~The reformulator model is only required to reformulate questions rather than solve them, thereby imposing lower demands on its reasoning capabilities.
To assess the generality of MQR to reformulator models with less capability, we utilize two smaller and open-source models: Qwen2.5-7B-Instruct \citep{qwen2025qwen2} and Qwen3-30B-A3B-Thinking \citep{yang2025qwen3}.
As shown in Table~\ref{tab:reformulator}, while the most capable OpenAI o3 reformulator achieves the best results, the other two models also deliver substantial gains over the original data.
This indicates that even moderately capable models can effectively generate challenging question reformulations that enhance mathematical reasoning within the MQR strategy.

\begin{table}[t!]
    \vspace{-0.05in}
    \caption{Comparative results of MQR using varying reformulator models on the MATH dataset.}
    \label{tab:reformulator}
    \vspace{-0.02in}
    \begin{center}
    \scalebox{0.88}{
    \tabcolsep3.9pt
    {\renewcommand{\arraystretch}{1.2}
        \begin{tabular}{l|cccccc|l}
        \toprule[1.2pt]
        Reformulators & AIME24 & AIME25 & AMC23 & MATH500 & Minerva & Olympiad & Avg./$\Delta_\text{Ori.}$ \\
        \midrule
        Original & 26.46 & 9.17 & 58.67 & 74.65 & 31.62 & 38.81 & ~~39.90 \\
        Qwen2.5-7B-Instruct & \underline{25.10} & 11.98 & 58.67 & 76.85 & 33.00 & 40.96 & ~~41.09\tiny{$\textcolor{up}{+1.19}$} \\
        Qwen3-30B-A3B-Thinking & \textbf{25.73} & \underline{12.29} & \textbf{59.84} & \underline{78.85} & \underline{33.18} & \underline{41.22} & ~~\underline{41.85}\tiny{$\textcolor{up}{+1.95}$} \\
        \rowcolor{lightblue} OpenAI o3 & 24.58 & \textbf{12.60} & \textbf{59.84} & \textbf{79.95} & \textbf{33.36} & \textbf{42.67} & ~~\textbf{42.17}\tiny{$\textcolor{up}{+2.27}$} \\
        \bottomrule[1.2pt]
        \end{tabular}}}
    \end{center}
    \vspace{-0.1in}
\end{table}

\vspace{-0.05in}
\section{Related Work}
\vspace{-0.05in}

\textbf{Reinforcement Learning.}~~Policy optimization has become a standard for post-training large language models to enhance their reasoning capabilities~\citep{jaech2024openai,guo2025deepseek,team2025kimi}. 
Building upon Proximal Policy Optimization (PPO) \citep{schulman2017proximal}, Group Relative Policy Optimization (GRPO) \citep{shao2024deepseekmath} proposes a highly efficient critic-less paradigm using group relative advantage estimation.
This spurred a line of research focused on refining GRPO's stability and performance.
For example, Dr.GRPO \citep{liu2025understanding} removes the length bias and PPO-objective bias in GRPO's advantage estimation.
GPG \citep{chu2025gpg}, DAPO \citep{yu2025dapo}, and GRPO-LEAD \citep{zhang2025grpo} address issues in reward design, advantage estimation, and oversampling, while GSPO \citep{zheng2025group} and GMPO \citep{zhao2025geometric} introduce alternative optimization objectives.
Besides, another line of work \citep{dai2025s,yue2025vapo,liu2025ghpo} proposes more complex pipelines, such as value models or prompt refinement.
RL for reasoning has been widely explored and adopted across diverse applications \citep{sun2025detection, ji2025tree, ji2026thinking, yang2025deepcritic, yang2025laser, li2025adacurl, xiong2025hs}.

\textbf{Data Augmentation.}~~A parallel line of work improves mathematical reasoning from a data-centric perspective. 
One strategy involves generating entirely new, high-quality problem-solution pairs using powerful teacher models, showing that synthetic data can rival human-curated datasets~\citep{luo2023wizardmath,li2024mugglemath,li2024common}. 
Another strategy, more aligned with our work, focuses on reformulating existing questions while preserving the original answer. 
Approaches like MetaMath~\citep{yu2024metamath} and PersonaMath~\citep{luo2024personamath} achieve this by rephrasing problems or adopting specific personas. Moreover, an advanced approach employs self-play, where the model generates its own challenging questions from solutions, fostering continuous self-improvement~\citep{liang2025beyond}.

\vspace{-0.05in}
\section{Conclusion}
\vspace{-0.05in}

In this paper, we propose MathForge, a comprehensive framework designed to enhance mathematical reasoning by targeting harder problems from both algorithmic and data perspectives.
MathForge is two-fold: the Difficulty-Aware Group Policy Optimization (DGPO) algorithm rectifies the update magnitude imbalance and prioritizes challenging questions, while the Multi-Aspect Question Reformulation (MQR) strategy augments training data with more difficult, yet answer-preserving, question variants from multiple aspects.
Extensive experiments demonstrate that this synergistic combination significantly outperforms existing methods across various models and benchmarks, underscoring our core principle that ``harder is better" in mathematical reasoning.

\vspace{-0.05in}
\section*{Acknowledgments}
\vspace{-0.05in}

This work was supported in part by National Natural Science Foundation of China (62376274, 62437002).

\section*{Ethics Statement}

This work adheres to the ICLR Code of Ethics, ensuring ethical compliance throughout all stages of the research.
The MQR-augmented data was constructed by reformulating problems from the public MATH dataset. This process and the source data do not involve any personally identifiable information or sensitive content, thereby mitigating privacy concerns.
The primary goal of our research is to enhance the mathematical reasoning capabilities of AI models, a pursuit with significant potential benefits for scientific research, engineering, and education.

\section*{Reproducibility Statement}

To ensure the full reproducibility of our research, we will make our code and the MQR-augmented dataset publicly available. 
Our implementation is built upon the Open-R1 codebase \citep{openr1}. 
Comprehensive details regarding the experimental setup, including model configurations and all hyperparameters, are described in Section~\ref{sec:setup} and further elaborated in Appendix~\ref{sec:appendix_implementation}. 
For the MQR strategy, the exact prompts used for generating the augmented data are provided in Appendix~\ref{sec:appendix_prompts}, and illustrative examples of the reformulated questions are presented in Appendix~\ref{sec:appendix_data}.

\bibliography{iclr2026_conference}
\bibliographystyle{iclr2026_conference}

\appendix

\section{The Use of Large Language Models (LLMs)}

We used LLMs to assist in polishing the writing of this paper. Its use was limited to improving grammar, clarity, and style. 
All core intellectual contributions, including the proposed methods, experimental design, and analysis, were conceived and executed by the human authors.

\section{Proofs}

\subsection{Full Derivation for Gradient of GRPO}\label{sec:appendix_gradient}

Consider a single question $q$ and its corresponding responses $\left\{o_i\right\}^G_{i=1}$, the unclipped policy gradient calculated in GRPO is as follows:
\begin{align}\label{eq:gradientpf}
    g_\text{GRPO} &= \frac{1}{\sum^G_{i=1}\left|o_i\right|}\sum^G_{i=1}\sum^{\left|o_i\right|}_{t=1}\hat{A}_{\text{GR},i}\nabla_\theta I_{it}(\theta) \nonumber\\
    &= \frac{1}{\sum^G_{i=1}\left|o_i\right|}\sum^G_{i=1}\sum^{\left|o_i\right|}_{t=1}\hat{A}_{\text{GR},i}\nabla_\theta \frac{\pi_\theta\left(o_{i,t}\mid q,o_{i,<t}\right)}{\pi_{\theta_\text{old}}\left(o_{i,t}\mid q,o_{i,<t}\right)} \nonumber \\
    &= \frac{1}{\sum^G_{i=1}\left|o_i\right|}\sum^G_{i=1}\sum^{\left|o_i\right|}_{t=1}\hat{A}_{\text{GR},i}\frac{\operatorname{detach}\left(\pi_\theta\left(o_{i,t}\mid q,o_{i,<t}\right)\right)}{\pi_{\theta_\text{old}}\left(o_{i,t}\mid q,o_{i,<t}\right)}\nabla_\theta \frac{\pi_\theta\left(o_{i,t}\mid q,o_{i,<t}\right)}{\operatorname{detach}\left(\pi_{\theta}\left(o_{i,t}\mid q,o_{i,<t}\right)\right)} \nonumber \\
    &= \frac{1}{\sum^G_{i=1}\left|o_i\right|}\sum^G_{i=1}\sum^{\left|o_i\right|}_{t=1}\hat{A}_{\text{GR},i}\operatorname{detach}\left(I_{it}(\theta)\right)\nabla_\theta\log\left(\pi_\theta\left(o_{i,t}\mid q,o_{i,<t}\right)\right) \nonumber \\
    &= \frac{1}{\sum^G_{i=1}\left|o_i\right|}\sum^G_{i=1}\sum^{\left|o_i\right|}_{t=1}\operatorname{sgn}\left(\hat{A}_{\text{GR},i}\right)\left|\hat{A}_{\text{GR},i}\right|\operatorname{detach}\left(I_{it}(\theta)\right)\nabla_\theta\log\left(\pi_\theta\left(o_{i,t}\mid q,o_{i,<t}\right)\right),
\end{align}
where $\operatorname{sgn}(\cdot)$ is the sign function and $\operatorname{detach}(\cdot)$ is the stop-gradient operator.

\subsection{Full Derivation for the Total Update Magnitude of GRPO}\label{sec:appendix_magnitude}

The PPO/GRPO-style gradient for a fixed question $q$ can be written (ignoring token length difference, clipping, and importance sampling terms) as:

\begin{equation}
    g(q) = \frac{1}{G}\sum^G_{i=1} \hat{A}_i \nabla_\theta \log \pi_\theta\left(o_i\mid q\right) \triangleq \frac{1}{G}\sum^G_{i=1} \hat{A}_i g_i,
\end{equation}

By the triangle inequality, the gradient norm satisfies:

\begin{equation}
    \|g(q)\| = \left\|\frac{1}{G}\sum^G_{i=1} \hat{A}_i g_i\right\| \leq \frac{1}{G}\sum^G_{i=1}|\hat{A}_i|\|g_i\|.
\end{equation}

Since all gradients $\hat{A}_ig_i$ are generated from the same question and tend to together improve the policy on that specific query, their directions are positively correlated.
Such directional alignment implies limited mutual cancellation, causing the triangle inequality to be nearly tight.

Moreover, as all responses in a batch are sampled from the same policy with the same temperature and the same or similar math prompt, the variation in $\|g_i\|$ is typically much smaller than the variation in $|\hat{A}_i|$.
Under this mild assumption, $\sum_i|\hat{A}_i|$ serves as a tight upper bound and a faithful proxy for the question-level update strength, but is not an exact equality.

\subsection{Proof of Theorem~\ref{th:grae}}\label{sec:appendix_grae}

We provide a proof of Theorem~\ref{th:grae} (Update Magnitude for a Single Question using GRAE) below.

\begin{proof}
By definition, the total update magnitude is the sum of the absolute values of the advantages:
\begin{equation}
    \sum_{i=1}^G \left|\hat{A}_{\text{GR},i}\right| = \sum_{i=1}^G \left|\frac{r_i-\operatorname{mean}\left(\{r_i\}_{i=1}^G\right)}{\operatorname{std}\left(\{r_i\}_{i=1}^G\right)}\right| = \frac{\sum_{i=1}^G \left|r_i-\operatorname{mean}\left(\{r_i\}_{i=1}^G\right)\right|}{\operatorname{std}\left(\{r_i\}_{i=1}^G\right)}.
\end{equation}
For binary rewards $r_i \in \{0,1\}$, the mean value is the accuracy rate $p = \frac{1}{G}\sum_{i=1}^G r_i$, and the standard deviation is $\sqrt{p(1-p)}$. Substituting these gives:
\begin{equation}
    \sum_{i=1}^G \left|\hat{A}_{\text{GR},i}\right| = \frac{\sum_{i=1}^G |r_i - p|}{\sqrt{p(1-p)}}.
\end{equation}
The numerator can be decomposed based on the reward values. There are $Gp$ terms where $r_i=1$ and $G(1-p)$ terms where $r_i=0$. Therefore:
\begin{align}
    \sum_{i=1}^G \left|\hat{A}_{\text{GR},i}\right| &= \frac{Gp|1-p| + G(1-p)|0-p|}{\sqrt{p(1-p)}} \nonumber\\
    &= \frac{Gp(1-p) + G(1-p)p}{\sqrt{p(1-p)}} \quad (\text{since}~p \in (0,1)) \nonumber\\
    &= \frac{2Gp(1-p)}{\sqrt{p(1-p)}} \nonumber\\
    &= 2G\sqrt{p(1-p)}.
\end{align}
\end{proof}

\subsection{Proof of Theorem~\ref{th:dgae}}\label{sec:appendix_dgae}

We provide a proof of Theorem~\ref{th:dgae} (Update Magnitude for a Single Question using DGAE) below.

\begin{proof}
By definition, the total update magnitude is the sum of the absolute values of the advantages:
\begin{equation}
    \sum_{i=1}^G \left|\hat{A}_{\text{DG},i}\right| = \sum_{i=1}^G \left|\frac{r_i-\operatorname{mean}\left(\{r_i\}_{i=1}^G\right)}{\frac{1}{G}\sum_{i=1}^G\left|r_i-\operatorname{mean}\left(\{r_i\}_{i=1}^G\right)\right|}\right|.
\end{equation}
Since the denominator, $\frac{1}{G}\sum_{j=1}^G \left|r_j - \operatorname{mean}\left(\{r_i\}_{i=1}^G\right)\right|$, is constant with respect to the summation index $i$ and non-negative, we can move it outside the outer summation:
\begin{equation}
    \sum_{i=1}^G \left|\hat{A}_{\text{DG},i}\right| = \frac{\sum_{i=1}^G \left|r_i - \operatorname{mean}(\{r_i\}_{i=1}^G)\right|}{\frac{1}{G}\sum_{i=1}^G \left|r_i - \operatorname{mean}(\{r_i\}_{i=1}^G)\right|} = G.
\end{equation}
\end{proof}

\section{Prompts for MQR}\label{sec:appendix_prompts}

We provide the detailed prompts for MQR below.

\begin{promptbox}{General Prompt for Question Reformulation}
I want you to act as an expert Math Question Rephraser.\\\\
Your goal is to rephrase a given math question so it becomes more challenging for large AI models while remaining logically sound and fully comprehensible to humans. The rephrased question MUST yield exactly the same final answer as the original.\\\\
You should complicate the given question using the following method:\\
\{instruction\}\\\\
You must strictly adhere to the following constraints:\\
- The final answer MUST remain unchanged.\\
- The rephrased question should be no more than 100 words longer than the given question.\\
- Preserve the original interrogative verb (e.g., “find”, “determine”, “compute…”, “evaluate”).\\
- Use LaTeX for all mathematical expressions.\\
- Output only the rephrased question (no hints, solutions, explanation, or commentary).\\\\
\#Given Question Start\#\\
\{question\}\\
\#Given Question End\#
\end{promptbox}

\begin{promptbox}{Specific Instruction for Background Question}
- Add a story background that is not related to the core mathematical content of the given question, but seems to be related to the question.\\
- If the given question already has such a background, change it to a new, complexer background.\\
- Possible background themes include, but are not limited to, the following: history, culture, geography, nature, occupation, daily life, sports, art, science fiction, and adventure. Astronomy is explicitly excluded.\\
- The background should be presented as natural parts of the question statement, ensuring the rephrased question is coherent and self-contained.
\end{promptbox}

\begin{promptbox}{Specific Instruction for Term Question}
- Invent a new, abstract mathematical term to define a concept that is central to the given question, and restate the entire question using this term.\\
- The term should be presented as natural parts of the question statement, ensuring the rephrased question is coherent and self-contained.
\end{promptbox}

\begin{promptbox}{Specific Instruction for Sub-Problem Question}
- Convert a key numerical condition of the given question which have a definite value into an independent sub-problem.\\
- The sub-problem may belong to any branch of mathematics (e.g., algebra, geometry, number theory, combinatorics).\\
- The sub-problem must be self-contained, have a unique solution, and its solution must yield exactly the value required for the original question.\\
- The sub-problem should be presented as natural parts of the question statement, ensuring the rephrased question is coherent and self-contained.
\end{promptbox}

\section{Augmented Data of MQR}\label{sec:appendix_data}

We provide examples of questions generated by MQR below, with the highlighted parts representing the main modifications made according to the reformulation strategies.

\begin{promptbox}{Original Question}
Berengere and her American foreign-exchange student Emily are at a bakery in Paris that accepts both euros and American dollars. They want to buy a cake, but neither of them has enough money. If the cake costs 6 euros and Emily has an American five-dollar bill, how many euros does Berengere need to contribute to the cost of the cake if 1 euro = 1.25 USD?
\end{promptbox}

\begin{promptbox}{Question using Background Reformulation}
\textcolor{blue}{In the bustling Montmartre district of Paris, Berengere—a culinary historian compiling notes on classic French desserts—and her visiting American friend Emily, an anthropology student documenting European food customs, wander into the venerable pâtisserie “Le Temps Sucré.”} They decide to purchase a famed gâteau Saint-Honoré that the proprietor has priced at \(6\) euros. Emily searches her travel wallet and discovers only a single crisp five-dollar bill, while Berengere carries euros exclusively. A sign by the register lists the day’s exchange rate as \(1\text{ euro}=1.25\text{ USD}\). To complete the purchase, how many euros must Berengere contribute?
\end{promptbox}

\begin{promptbox}{Question using Term Reformulation}
\textcolor{blue}{Define the “euro-gap” \( \epsilon \) of a prospective purchase as the non-negative difference, measured in euros, between an item’s listed euro price and the euro-denominated value of the funds already on hand to pay for it.} Berengere and her American foreign-exchange student Emily visit a Parisian bakery. The cake they wish to buy is priced at \(6\) euros. Emily can contribute only an American five-dollar bill, and the prevailing conversion rate is \(1\text{ euro}=1.25\text{ USD}\). \textcolor{blue}{Determine, in euros, the euro-gap \( \epsilon \) that Berengere must cover to complete the purchase.}
\end{promptbox}

\begin{promptbox}{Question using Sub-Problem Reformulation}
Berengere and her American foreign-exchange student Emily are at a Paris bakery that accepts both euros and U.S. dollars, but neither of them alone can pay for the desired cake.  \textcolor{blue}{Before the exchange rate is revealed, solve this independent task:  Find positive integers $x$ and $y$ that satisfy  \[x+y=9\quad\text{and}\quad x^{2}+y^{2}=41.\]  Let $r$ be the ratio of the larger of $x$ and $y$ to the smaller.The cashier states that €1 is worth exactly $r$ U.S. dollars.} The cake costs €6, and Emily offers a single \$5 bill.  Using the exchange rate $r$ defined above, how many euros must Berengere contribute so that together they can pay for the cake?
\end{promptbox}

\section{Data Augmentation Cost of MQR}\label{sec:appendix_cost}

The average token usage per question is 255.05 input tokens, 820.27 output reasoning tokens, and 138.33 output reformulated question tokens. Therefore, the total cost for generating 22.5k reformulated questions of the MATH dataset is approximately \$184.

\section{Implementation Details}\label{sec:appendix_implementation}

This section provides detailed information on the training and evaluation configurations used in our experiments.

For all reinforcement learning experiments, responses were generated with a temperature of $1.0$ and a maximum completion length of $1024$ tokens. During evaluation, we used a generation temperature of $0.6$, a top-p value of $0.95$, and set the maximum new tokens to $4096$.

\subsection{MATH}

For experiments trained on the MATH dataset, we used the following system prompt to guide the model's reasoning process: ``Please reason step by step, and put your final answer within \textbackslash boxed\{\}.''
The maximum prompt length was set to $512$ tokens. For each prompt, we generated $8$ responses and used a training batch size of $32$. 
The reward was based on binary accuracy, where a correct final answer yielded a reward of $1$ and an incorrect one yielded $0$.

Model-specific hyperparameters, including learning rate, number of epochs, gradient accumulation steps, and total training steps, are detailed in Table~\ref{tab:parameters}. The table specifies configurations for training on both the original 7.5k MATH dataset and the 30k MQR-augmented dataset.

\begin{table}[h!]
    \vspace{-0.15in}
    \caption{Hyperparameter settings trained on the MATH dataset using varying base models.}
    \label{tab:parameters}
    \begin{center}
    \scalebox{0.88}{
    \tabcolsep8pt
    {\renewcommand{\arraystretch}{1.2}
        \begin{tabular}{l|ccccccc}
        \toprule[1.2pt]
        \multirow{2}{*}{Models} & Learning & \multirow{2}{*}{Epochs} & Gradient & Training \\
        \multirow{2}{*}{} & Rate & \multirow{2}{*}{} & Accumulation & Steps \\
        \midrule
        Qwen2.5-Math-7B & 5e-7 & 1 & 1 & 230 \\
        ~+MQR & 1e-6 & 1 & 4 & 230 \\
        \midrule
        Qwen2.5-Math-1.5B & 5e-7 & 1 & 1 & 230 \\
        ~+MQR & 1e-6 & 1 & 4 & 230 \\
        \midrule
        Qwen2.5-3B & 5e-7 & 1 & 1 & 230 \\
        ~+MQR & 1e-6 & 1 & 4 & 230 \\
        \midrule
        DeepSeek-Math-7B & 1e-6 & 2 & 1 & 468 \\
        ~+MQR & 1e-6 & 1 & 1 & 937 \\
        \bottomrule[1.2pt]
        \end{tabular}}}
    \end{center}
    \vspace{-0.1in}
\end{table}

For the cold start of DeepSeek-Math-7B, we sampled 80k data from NuminaMath-CoT to fine-tune it with a learning rate of $2e{-6}$, a batch size of $32$, and gradient accumulation steps of $8$, resulting in a total of $40$ training steps.

\subsection{GEOQA-8k}

For the multimodal experiments on the GEOQA-8k dataset using Qwen2.5-VL-3B-Instruct, we performed a preprocessing step to remove non-standard units from the gold answers to facilitate consistent reward calculation. 
Consequently, the system prompt was adjusted to: ``Please reason step by step, and put your final answer without units in \textbackslash boxed\{\}.''

The training was configured with a maximum prompt length of $2048$ tokens and $8$ generated responses per question. 
The model was trained for $2$ epochs using a learning rate of $1e{-6}$ and a batch size of $32$. 
We set gradient accumulation steps to $1$, resulting in a total of $480$ training steps. 
The reward mechanism was the same binary accuracy metric used in the text-only experiments.

\section{Combination Forms of DGPO and Other Methods}\label{sec:appendix_combination}

This section details how DGPO is integrated with other policy optimization methods.

\subsection{GPG}

The integration with GPG involves replacing its original advantage formulation with our DGAE and incorporating the DQW scheme. 
Specifically, the policy gradient objective of GPG is retained, but the update for each token is now scaled by the difficulty-balanced advantage $\hat{A}_{\text{DG},si}$. Furthermore, the loss contribution of each question is modulated by the difficulty-aware weight $\lambda_s$. The normalization is also adjusted to average over valid tokens. 
The optimization objective is as follows:
\begin{multline}
\mathcal{J}_\text{GPG+DGPO}(\theta)=\mathbb{E}\left[\left\{q_s\right\}^B_{s=1}\sim\mathcal{D},\left\{o_{si}\right\}^G_{i=1}\sim\pi_{\theta}(\cdot\mid q_s)\right]\\
\frac{1}{\sum^{\textcolor{blue}{B_\text{v}}}_{s=1}\sum^G_{i=1}\left|o_{si}\right|}\sum^{\textcolor{blue}{B_\text{v}}}_{s=1}\textcolor{blue}{\lambda_s}\sum^G_{i=1}\sum^{\left|o_{si}\right|}_{t=1}\left[-\log \pi_\theta\left(o_{i,t}\mid q,o_{i,<t}\right)\textcolor{blue}{\hat{A}_{\text{DG},si}}\right],
\end{multline}
where $\hat{A}_{\text{DG},si}$ is the advantage of the response $o_i$ obtained by DGAE given by:
\begin{equation}
    \hat{A}_{\text{DG},si}=\frac{r_{si}-\operatorname{mean}\left(\left\{r_{si}\right\}^G_{i=1}\right)}{\operatorname{MAD}\left(\left\{r_{si}\right\}^G_{i=1}\right)},
\end{equation}
and $\lambda_s$ is the difficulty-aware weight for the query $q_s$ computed by DQW as follows:
\begin{equation}
    \lambda_s = B_\text{v}\cdot\frac{\exp\left(D_s/T\right)}{\sum^{B_\text{v}}_{s=1}\exp\left(D_s/T\right)},~~\text{where}~D_s=-\operatorname{mean}\left(\left\{r_{si}\right\}^G_{i=1}\right).
\end{equation}

\subsection{DAPO}

For DAPO, the combination preserves its core PPO-style clipped objective and its use of a composite reward signal (accuracy plus length penalty, i.e., $r_{si} = r_{\text{acc},si} + r_{\text{length},si}$). We replace DAPO's original advantage estimation with our DGAE ($\hat{A}_{\text{DG},si}$), which is calculated using this composite reward. Crucially, the difficulty score $D_s$ for our DQW scheme is computed only using the accuracy component of the reward ($r_{\text{acc},si}$). 
This design choice ensures that the question weighting focuses purely on the logical difficulty of the question, rather than being conflated with the verbosity of the responses. 
The optimization objective is as follows:
\begin{multline}
\mathcal{J}_\text{DAPO+DGPO}(\theta)=\mathbb{E}\left[\left\{q_s\right\}^B_{s=1}\sim\mathcal{D},\left\{o_{si}\right\}^G_{i=1}\sim\pi_{\theta}(\cdot\mid q_s)\right]\\
\frac{1}{\sum^{\textcolor{blue}{B_\text{v}}}_{s=1}\sum^G_{i=1}\left|o_{si}\right|}\sum^{\textcolor{blue}{B_\text{v}}}_{s=1}\textcolor{blue}{\lambda_s}\sum^G_{i=1}\sum^{\left|o_{si}\right|}_{t=1}\left\{\min\left[I_{sit}(\theta)\textcolor{blue}{\hat{A}_{\text{DG},si}},\operatorname{clip}\left(I_{sit}(\theta),1-\varepsilon_\text{low},1+\varepsilon_\text{high}\right)\textcolor{blue}{\hat{A}_{\text{DG},si}}\right]\right\},
\end{multline}
where $I_{sit}(\theta)$ is the importance sampling ratio of the token $o_{si,t}$, and $\hat{A}_{\text{DG},si}$ is the advantage of the response $o_i$ obtained by DGAE, respectively given by:
\begin{equation}
    I_{sit}(\theta)=\frac{\pi_\theta\left(o_{si,t}\mid q_s,o_{si,<t}\right)}{\pi_{\theta_\text{old}}\left(o_{si,t}\mid q_s,o_{si,<t}\right)},~~\hat{A}_{\text{DG},si}=\frac{r_{si}-\operatorname{mean}\left(\left\{r_{si}\right\}^G_{i=1}\right)}{\operatorname{MAD}\left(\left\{r_{si}\right\}^G_{i=1}\right)},
\end{equation}
and $\lambda_s$ is the difficulty-aware weight for the query $q_s$ computed by DQW as follows:
\begin{gather}
    \lambda_s = B_\text{v}\cdot\frac{\exp\left(D_s/T\right)}{\sum^{B_\text{v}}_{s=1}\exp\left(D_s/T\right)}, \nonumber\\
    \text{where}~D_s = \begin{cases} -\operatorname{mean}\left(\left\{r_{\text{acc},si}\right\}^G_{i=1}\right) & \text{if}~\operatorname{mean}\left(\left\{r_{\text{acc},si}\right\}^G_{i=1}\right) \neq 0 \\ -1 & \text{if}~\operatorname{mean}\left(\left\{r_{\text{acc},si}\right\}^G_{i=1}\right) = 0 \end{cases}.
\end{gather}
Here, $B_\text{v}$ signifies the number of valid queries in the batch.
A query is considered valid if its rewards for $G$ corresponding responses are not completely equal.
For questions where all corresponding responses are incorrect (i.e., accuracy reward is $0$), no positive learning signal is available in the current question. 
Consequently, we deliberately set its corresponding difficulty score, $D_s$, to its floor value of $-1$. 
This prevents the model from dedicating excessive attention to instances that offer no constructive gradient for policy improvement.

\subsection{GSPO}

The integration with GSPO is performed at the sequence level, aligning with GSPO's fundamental design. GSPO's sequence-level importance sampling ratio ($S_{si}$) is preserved. The update for each sequence is then driven by our DGAE, $\hat{A}_{\text{DG},si}$. The question-level weighting $\lambda_s$ is also applied to modulate the influence of each question on the total loss. The loss is averaged over the number of valid questions, which aligns with the sequence-level nature of both GSPO and our DGPO.
The optimization objective is as follows:
\begin{multline}
\mathcal{J}_\text{GSPO+DGPO}(\theta)=\mathbb{E}\left[\left\{q_s\right\}^B_{s=1}\sim\mathcal{D},\left\{o_{si}\right\}^G_{i=1}\sim\pi_{\theta}(\cdot\mid q_s)\right]\\
\frac{1}{\textcolor{blue}{B_\text{v}}\cdot G}\sum^{\textcolor{blue}{B_\text{v}}}_{s=1}\textcolor{blue}{\lambda_s}\sum^G_{i=1}\left\{\min\left[S_{si}(\theta)\textcolor{blue}{\hat{A}_{\text{DG},si}},\operatorname{clip}\left(S_{si}(\theta),1-\varepsilon,1+\varepsilon\right)\textcolor{blue}{\hat{A}_{\text{DG},si}}\right]\right\},
\end{multline}
where $S_{si}(\theta)$ is the sequence-level importance sampling ratio of the response $o_{si}$, and $\hat{A}_{\text{DG},si}$ is the advantage of the response $o_i$ obtained by DGAE, respectively given by:
\begin{equation}
    S_{si}(\theta)=\left(\prod^{|o_{si}|}_{t=1}\frac{\pi_\theta\left(o_{si,t}\mid q_s,o_{si,<t}\right)}{\pi_{\theta_\text{old}}\left(o_{si,t}\mid q_s,o_{si,<t}\right)}\right)^{\frac{1}{|o_{si}|}},~~\hat{A}_{\text{DG},si}=\frac{r_{si}-\operatorname{mean}\left(\left\{r_{si}\right\}^G_{i=1}\right)}{\operatorname{MAD}\left(\left\{r_{si}\right\}^G_{i=1}\right)},
\end{equation}
and $\lambda_s$ is the difficulty-aware weight for the query $q_s$ computed by DQW as follows:
\begin{equation}
    \lambda_s = B_\text{v}\cdot\frac{\exp\left(D_s/T\right)}{\sum^{B_\text{v}}_{s=1}\exp\left(D_s/T\right)},~~\text{where}~D_s=-\operatorname{mean}\left(\left\{r_{si}\right\}^G_{i=1}\right).
\end{equation}

\section{Quality Assessment of MQR}\label{sec:appendix_quality}

We utilized the OpenAI o3 model to determine whether a reformulated question is mathematically equivalent to the original question. 
In this context, mathematical equivalence is defined as the capacity to yield the same final answer. 
The specific prompt used for this evaluation is as follows:

\begin{promptbox}{Prompt for Quality Assessment of MQR}
You are an expert in mathematics and logic.\\\\
Your task is to meticulously analyze and compare two versions of a mathematical problem: an ``Original Question" and a ``Rewritten Question". Your primary objective is to determine if these two questions are mathematically equivalent. For the purpose of this task, "mathematically equivalent" means that both questions, when solved correctly, will yield the identical final numerical answer or symbolic solution.\\\\
Please structure your response as follows: 1. \textbf{Equivalence Verdict:} Start with a clear and unambiguous ``Yes" or ``No". 2. \textbf{Detailed Justification:} If they are equivalent, explain why the changes in wording, structure, or given information do not alter the underlying mathematical operations or the final result. If they are not equivalent, pinpoint the specific change in the rewritten question that alters the problem's mathematical core. Explain how this change leads to a different solution or answer.\\\\
\#Original Question Start\#\\
\{question\}\\
\#Original Question End\#\\\\
\#Rewritten Question Start\#\\
\{rewritten\_question\}\\
\#Rewritten Question End\#
\end{promptbox}

We randomly sampled 100 questions from each of the three categories of reformulated questions, which yielded equivalence rates of 99\% for Background, 97\% for Term, and 97\% for Sub-Problem, respectively.

In MQR, a failed reformulation means that the resulting question becomes unsolvable or has a new answer different from the original answer.
In math reasoning RLVR, the answer space is open-ended, extremely large, and requires exact canonical matching (e.g., exact integers, simplified fractions, or normalized symbolic expressions).
Therefore, it is highly improbable that the policy model would reason incorrectly and happen to provide the same answer as that of the original question. Therefore, the multiple responses to the corrupted question would be uniformly incorrect (i.e., all rewards = 0).
Under GRPO and its variants (including our DGPO), such questions are invalid queries yielding no update gradients, thereby providing no harmful training signals.

\end{document}